\documentclass{IET}%%%%where IET is the template name

\usepackage{color}
\usepackage{threeparttable}
\usepackage{multirow}
\usepackage{booktabs,tabularx}

\usepackage{amsmath}

\graphicspath{{./figures/}}

\newcommand{\etal}{et al.~}

\begin{document}

\title{Face Deidentification with Generative Deep Neural Networks}
% Face De-identification using Generative Neural Networks

\author[1*]{Bla\v{z} Meden}
\affil{Faculty of Computer and Information Science, University of Ljubljana, Ve\v{c}na pot 113, SI-1001 Ljubljana, Slovenia}

\author[2]{Refik Can Mall{\i}}
%%%% By default, the citations will come automatically,
%%%% The optional bracket "[2.*]" is used  to display the corresponding author symbol
\affil{Department of Computer Engineering, Istanbul Technical University, 34469 Maslak, Istanbul, Turkey}

\author[3]{Sebastjan Fabijan}
\affil{Faculty of Electrical Engineering, University of Ljubljana, Tr\v{z}a\v{s}ka cesta 25, SI-1000 Ljubljana, Slovenia}

\author[2]{Haz{\i}m Kemal Ekenel}
\author[3]{Vitomir \v{S}truc}
\author[1]{Peter Peer}
%%%% Corresponding author detail must placed here
\affil[*]{blaz.meden@fri.uni-lj.si}

\abstract{Face deidentification is an active topic amongst privacy and security researchers. Early deidentification methods relying on image blurring or pixelization were replaced in recent years with techniques based on formal anonymity models that provide privacy  guaranties and at the same time aim at retaining certain characteristics of the data even after deidentification. The latter aspect is particularly important, as it allows to exploit the deidentified data in applications for which identity information is irrelevant. In this work we present a novel face deidentification pipeline, which ensures anonymity by synthesizing artificial surrogate faces using generative neural networks (GNNs). The generated faces are used to deidentify subjects in images or video, while preserving non-identity-related aspects of the data and consequently enabling data utilization. Since generative networks are very adaptive and can utilize a diverse set of parameters (pertaining to the appearance of the generated output in terms of facial expressions, gender, race, etc.), they represent a natural choice for the problem of face deidentification. To demonstrate the feasibility of our approach, we perform experiments using automated recognition tools and human annotators. Our results show that the recognition performance on deidentified images is close to chance, suggesting that the deidentification process based on GNNs is highly effective. %We conclude our research by indicating that a deep generative model can be used in the deidentification process. 
%We observed that  in order to improve the performance of the proposed pipeline further, such as support of non-frontal facial deidentification or advanced skin-tone corrections should be more thoroughly explored
}

\maketitle
 
\makeatletter
\def\blfootnote{\xdef\@thefnmark{}\@footnotetext}
\makeatother

\blfootnote{This paper is a postprint of a paper submitted to and accepted for publication in IET Signal Processing, Special Issue on Deidentification on May 3, 2017 and is subject to Institution of Engineering and Technology Copyright. The copy of record is available at IET Digital Library}
%\lhead{Guides and tutorials}

\section{Introduction}
%\textcolor{red}{
%TO WRITE: Describe the problem and existing solutions. Identify shortcomings and introduce possible solutions that coincide with our pipeline. Motivate.
%Say what we have done and summarize our main findings and observations. Compare our findings with what is possible with existing models.
%}
Over the last decades an extensive amount of video data is being recorded and stored. Since access to and exploitation of such data is difficult to monitor let alone prevent, appropriate measures need to be taken to ensure that such data is not misused and the privacy of people is adequately protected.

%Over the last decades an extensive amount of video data is being recorded and stored.
%It is of paramount importance to take appropriate measures  to ensure that such data is not misused and the privacy of people is adequately protected. 

%Currently there are no formal requirements (such as legislation, agreements and alike) in place, which would ensure that the data held or collected by any party is adequatly anonymized before being shared to or accessed by someone else.

%Over the last decades an extensive amount of video data is being recorded and stored. Since accessing, processing, and exploitation of such data is difficult to monitor let alone prevent, appropriate measures need to be taken to ensure that such data is not misused and the privacy of people is adequately protected. Currently there are no formal requirements (such as legislation, agreements and alike) in place, which would ensure that the data held or collected by any party is adequatly anonymized before being shared to or accessed by someone else.

A popular approach towards privacy protection in image and video data is the use of deidentification. Ribari\'{c} et al. \cite{Ribaric_Review2016} define deidentification as the process of concealing or removing personal identifiers from source content in order to prevent disclosure and use of data for unauthorized purposes. For video data, for example, this may translate to ``blurring'' or ``pixelation'' of the facial areas~\cite{PixelizationAndBlurring}, both of which represent early deidentification examples. These naive methods are typically useful for preventing humans from recognizing subjects in videos, but are far less successful with automated recognition techniques, where repeating the (naive) deidentification process on the test data still enables automated recognition, i.e., parrot attack~\cite{Newton_original}. Thus, for successful deidentification of images and videos, more advanced techniques are needed.

Another shortcoming of naive deidentification techniques is the fact that all information contained in the data is typically removed even if the information is not related to identity. This raises the question of data utility. If the deidentified data is to be useful for purposes that do not require identity information, but, for example, rely on gender or age information (e.g., customer-profiling applications in shopping malls), this information needs to be preserved even after deidentification. Recent deidentification approaches, therefore, focus on ways of removing identity information from images and videos, while still retaining other non-identity related information~\cite{Gross_utility},~\cite{Sim2015}.

In this paper we follow these recent trends and present a new deidentification approach exploiting generative neural networks (GNNs), which represent contemporary generative models capable of synthesizing photo-realistic artificial images of any object (see, e.g., \cite{Goodfellow_GAN2014}, \cite{Dosovitskiy_Chairs2015}, \cite{VAE_GAN}) based on supplied high-level information. Similarly to existing deidentification techniques, we replace the original faces in the input data with surrogates generated from a small number of identities. However, instead of synthesizing the surrogate faces through pixel averaging as in prior work, we use a GNN to combine identities and generate artificial surrogates for deidentification. The flexibility of the GNN also allows us to parameterize the generation process with respect to various appearance-related characteristics and synthesize faces under different appearances (under varying pose, with different facial expressions, etc.). This property ensures that our deidentification approach is able to conceal the identity of individuals, but also to preserve the utility of the data.

We demonstrate the feasibility of the proposed deidentification pipeline through extensive experiments on the ChokePoint dataset~\cite{Wong_Chokepoint2011}. Our experimental results show that GNNs are a viable solution for the problem of face deidentification and are able to generate realistic, visually convincing deidentification results. Furthermore, the deidentified faces offer a suitable level of privacy protection as evidenced by experiments with a number of contemporary recognition models as well as humans.   
%however the deidentification research field is very diverse and active in the last few years. Naive methods, such as blacking-out, blurring or pixelation are being replaced with more advanced methods, which in most cases can provide a formal anonymity model or are at least trying to be close to this concept. Such methods should be used worldwide, to automatically prevent extensive gathering of personal data and identity tracking in video surveillance systems -- a practical example is sharing the video data while preserving the privacy of recorded subjects.
%Additionally -- by using these methods -- the people appearances in videos would look more natural and thus could preserve more non-identity related properties, which could be still useful in cases, where forensic analysis can be performed on the captured data (analysis of expressed emotions on a face to gain more information about the incidents for example).
In summary, we make the following contributions: % in this paper:
\begin{itemize}
\item We introduce a face deidentification pipeline that exploits GNNs to produce artificial surrogate faces for deidentification and offers a level of flexibility in the generation process that  is not available with existing deidentification approaches.
\item We present a qualitative evaluation of the proposed pipeline with challenging data captured in a real surveillance scenario and discuss the advantages and limitations of our deidentification approach.
\item We demonstrate the efficacy of the proposed pipeline in comprehensive quantitative experiments with several state-of-the-art recognition techniques from the literature and human annotators.
\end{itemize}

%Recent methods are focused on Active Appearance Models, some of them try to separate identity properties from non-identity properties. 

%To the best of our knowledge, there is no published approach yet, which would use generative neural networks to perform the de-identification process on video data.

\section{Related work}

In this section we review the most important work related to our deidentification pipeline. %The main goal of this section is to position our work within the broader research field and point to the merits of our deidentification approach. 
For a more comprehensive review please refer to the surveys by Ribari\'{c} et al.~\cite{Ribaric_Review2016},~\cite{Ribaric_Re}.

%\subsection{De-identification approaches}

%Here we review the most important work related to de-identification of facial images, which also represents the focus of our work. 

Existing approaches to deidentification often implement formal privacy protection models such as $k$-anonymity~\cite{Sweeney_Kanonym}, $l$-diversity~\cite{Machana_Ldiversity}, or $t$-closeness~\cite{Li_Tcloseness}. Among these, the $k$-anonymity models have likely received the most attention in the area of face deidentification and resulted in the so-called $k$-same family of algorithms~\cite{Newton_original},~\cite{Gross_utility},~\cite{Gross_MFM2008}. These algorithms operate on a closed set of static facial images and substitute each image in the set with the average of the closest $k$ identities computed from the same closed set of images. Because several images are replaced with the same average face, data anonymity of a certain level is guaranteed. A number of $k$-same variants was presented in the literature, including the original $k$-same algorithm~\cite{Newton_original},  $k$-same-select~\cite{Gross_utility}, and  $k$-same-model~\cite{Gross_kSameM} to name a few. The majority of these techniques is implemented using Active Appearance Models (AAMs). %which represent the dominant tool for swapping faces in deidentification approaches.  

%\textcolor{red}{TO WRITE: most important papers, ideas and concepts. Add some images and equations if needed. Focus on important journal papers (IEEE, etc.), important vision conferences and recent years. The focus should be on the ideas and implementations (not necessarily the quantitative results).}

% TODO: some basic papers from Sweeney or Gross + AAM approaches

%Gross \etal proposed a multi-factor model-based deidentification approach in \cite{Gross_MFM2008} unifying linear, bilinear and quadratic models while enabling semi-supervised
%learning. The emphasis of this approach is the data utilization, where the facial expressions are preserved even if the $k$-same de-identification algorithm is applied to the subject images.

% AAM

Another example of a deidentification technique using AAMs was recently presented by Joura\-bloo et al. in \cite{Jourabloo_AAM2015}. Here, the authors combine facial-attribute and face-verification classifiers in a joint objective function. By optimizing the objective function, optimal weights are estimated such that the deidentified  and the original image have as many common attributes as possible, but at the same time are classified as two different subjects. %A second example of using AAMs for deidentification was presented by Du \etal in \cite{Du_GARP2014}. Here, a set of initial classifiers is first used to determine image attributes and then a deidentified face is generated using attribute specific AAMs.

A $q$-far deidentification approach, which is also AAM based, but does not follow the $k$-same principle (since the surrogate faces are mutually different), was proposed by Samar\v{z}ija and Ribari\'{c} in~\cite{Samarzija_2014} and combines face deidentification with pose estimation.  The authors cover different facial orientations by fitting multiple AAMs and achieve anonymity by replacing the original faces with surrogates that are sufficiently far (i.e., $q$-far) from the initial identities.

Sim and Zhang present a method for controllable face deidentification in \cite{Sim2015}. They demonstrate a high  degree of control over different attributes (such as identity, gender, age, femininity or race) of the deidentified faces and similar to our approach are able to alter or retain specific aspects of the target appearance. %The method operates by encoding the appearance and shape information of the input faces in compact vector forms. Next, a subspace decomposition using Multimodal Discriminant Analysis (MMDA) is performed, which enables controllable face alteration in accordance with predefined parameters. 

Another example related to the AAM-based techniques was recently proposed by Sun et al. \cite{Sun2015}. Here, the authors propose the $k$-diff-furthest algorithm, which is related to the $k$-anonymity model and $k$-same family of techniques, but differs in its ability to  track individuals in the deidentified video, since the deidentified faces have distinguishable properties. %This approach generates a unique deidentified face for each of the $k$ original faces in a cluster.
The experimental evaluation performed by the authors shows that the algorithm is capable of maintaining the diversity of the deidentified faces and keeps them as distinguishable as their original faces. However, the approach does not deal with the data utility aspect, e.g., expression preservation.

Different from AAM-based deidentification approaches, Brki\'{c} et al.~\cite{Brkic_ArtBased2016} propose a deidentification method based on style-transfer. The authors describe a pipeline that enables altering the appearance of faces in videos in such a way that an artistic style replacement is performed on the input data, thus making automatic recognition more difficult. Another interesting deidentification approach was presented by Chriskos et al. in \cite{Chriskos2015}, which, in contrary to most deidentification methods, hinders the recognition only for automatic recognition algorithms, but not human observers. The authors utilize projections on hyperspheres in order to defeat classifiers, while preserving enough visual information to enable human viewers to correctly identify individuals.

\section{Deidentification with generative deep neural networks}

Here we present a detailed description of our deidentification pipeline, which exploits generative neural networks (GNNs) to conceal the identity of people in the image data. %We start the section with a brief overview of our approach and then elaborate on the individual steps in the following sections.

\subsection{Overview}

\begin{figure}[htb]
\centering
\includegraphics[width=\columnwidth]{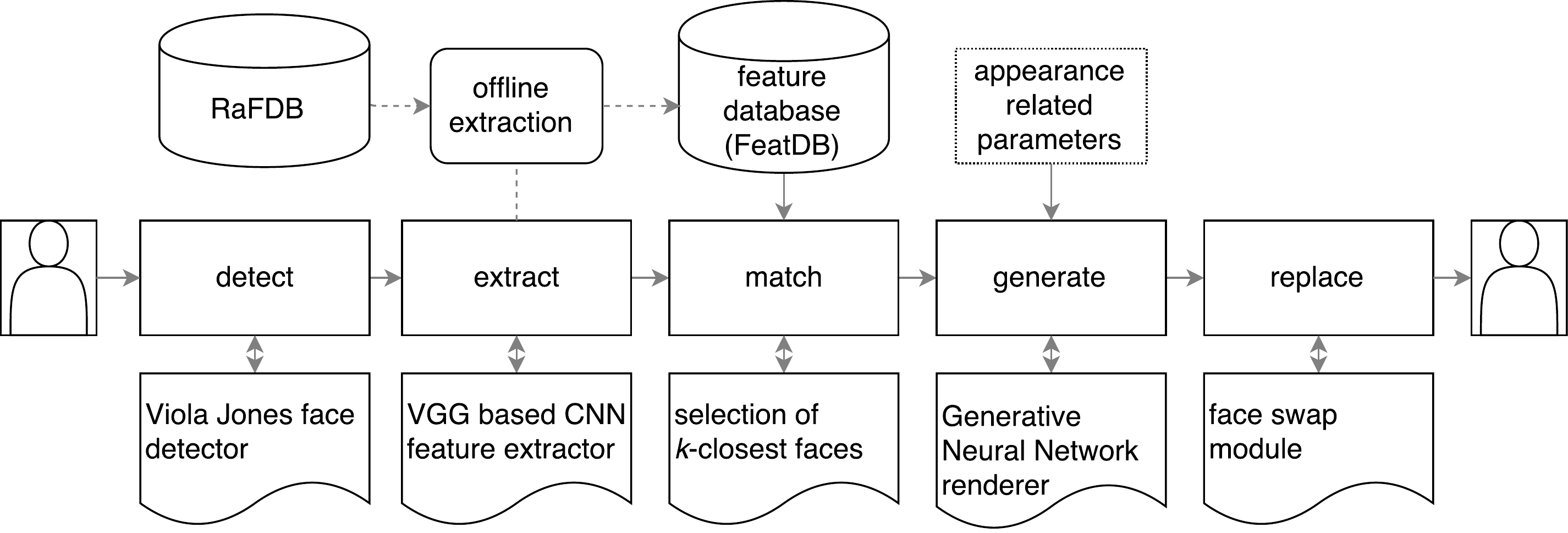}
\caption{Block diagram of our deidentification pipeline. The procedure uses a generative neural network to generate synthetic faces that can be used for deidentification. Each generated face is a combination of $k$ identities from the gallery data that are closest (i.e., most similar in the feature space) to the input face.}
\label{fig:pipe}
\end{figure}

A block-diagram of our deidentification pipeline is presented in Fig.~\ref{fig:pipe}. The procedure starts with a face detection step  that takes an image or video frame as input and then locates candidate facial regions in the input data for deidentification. For each detected region, a feature vector is computed using a state-of-the-art deep face recognition network, i.e., the VGG network from~\cite{Parkhi_Recog2015}, and matched against a fixed gallery of $M$ subjects. Based on this matching procedure, the $k$-closest identities ($k\ll M$) from the gallery data are selected and fed to our generative network to synthesize an artificial face with visual characteristics of the selected $k$ identities. Finally, the artificially generated (deidentified) face is blended into the input image (or frame) to conceal the original identity.      

%This step is shared with the $k$-same family of techniques that implement the $k$-anonymity model of
%\textcolor{red}{TO WRITE: General idea and overview of our pipeline. Motivate, motivate, motivate. A figure should be added here with a %block diagram of our approach.
%We also need arguments for the pipeline - what are the benefits compared to existing models based on AAMs. This could be a bulleted %list with explanations, e.g.:
%\begin{itemize}
%  \item Flexibility: bla bla
%  \item Adaptivity:
%  \item Another one:
%\end{itemize}
%}
	
One appealing characteristic of our deidentification pipeline is the flexibility of the GNN, which is able to synthesize high-resolution, realistic-looking faces under various appearances. Here, the generation process is governed by a small number of appearance-related parameters that control the visual characteristics of the synthesized faces, such as pose, skin color, gender, identity, facial expression, and alike. Thus, with this setup, we are able to generate artificial faces with predefined identities, facial expressions, gender, and so forth, or alter any of these at the time. For instance, if our goal is to preserve facial expressions of faces, we could automatically recognize facial expressions from the input image and use the recognition result as input to the GNN. The network would then generate a synthetic image with the predefined expression. A similar procedure could be used to retain or alter any visual characteristic of the input faces and contribute towards the preservation of data utility, which is one of the main goals of contemporary deidentification technology.

Even though our approach is similar in nature to the $k$-same family of algorithms~\cite{Newton_original},~\cite{Gross_utility},~\cite{Ribaric_Re} that implement the $k$-anonymity protection model~\cite{Sweeney_Kanonym}, there are important differences that invalidate some of the $k$-anonymity model assumptions. For example, our technique does not operate on a subject-specific set of images (with one image per subject only) nor is it limited to closed set scenarios. Thus, the anonymity guarantees associated with the $k$-same family of algorithms do not apply to our approach, so we use extensive experimental validation to demonstrate the feasibility of the developed deidentification pipeline. 

%Our pipeline architecture allows easy integration of additional pre- or post-processing blocks. In the following section describe the individual components of the pipeline, e.g., preprocessing, analysis, generation, blending (or other building blocks).

\subsection{Face detection and target identity estimation}
%%\subsection{Input}

%The input to our deidentification pipeline is either an RGB image or a sequence of images, in which case deidentification  is executed on a per-frame basis. %The pipeline  does not  exploit temporal information yet, as people are currently not being tracked but are only detected in each frame. However, future modifications could include the incorporation of tracking technology to improve the deidentification process in the case of video data. %, where the detector fails to detect all targets or when the target identity of deidentified subjects is not persistent throughout a videoswitches happens because of various environment effects (e.g. change of illumination on involved faces). The pipeline can also process images, where there are multiple people present in the scene.
%\subsection{Face Detection \label{sub:det}}
Our deidentification procedure starts with a standard face detection step using the off-the-shelf Viola-Jones face detector from OpenCV \cite{ViolaJones_Detector2001}. The detector process the input image or video frame and returns bounding boxes of all detected faces. Each detected region is then processed separately in a sequential manner and a 4096-dimensional feature vector is extracted from each region using the pre-trained 16-layer VGG face network from~\cite{Parkhi_Recog2015}. For this step, the output of the last fully-connected layer of the VGG face network is considered as a feature vector. Each computed feature vector is matched against a gallery of feature vectors using the cosine similarity.   %The detector uses Haar features to distinguish between positive and negative samples (e.g. to tell which sample represents a face or not) and a Cascade of Classifiers to optimize the detection process.

The matching procedure between the feature vector extracted from a region of the input image and the gallery of feature vectors results in an ordered list of similarity scores. Based on this list, we identify the $k$ most similar identities (where $k\ll M$) in our gallery and feed these to the generative network for surrogate-face generation. The idea of generating a synthetic surrogate face based on $k$ closest identities is similar in essence to the established family of $k$-same family of algorithms, except for the fact that the final face is in our case entirely generated by a GNN.

To generate the gallery for our deidentification approach, we process the images from the Radboud Faces Database (RaFDB) \cite{Langner_RAFDB2010} with the VGG network during an offline extraction step and store templates of all $M$ identities of the RaFDB dataset in the so-called feature database, FeatDB in Fig.~\ref{fig:pipe}, of our pipeline. These feature vectors correspond to a finite set of facial identities that can be used for generating new (surrogate) faces for deidentification.

\subsection{Face generation \label{sub:gen}}

The generative part used in our deidentification pipeline comprises a powerful GNN
%generative neural network 
recently introduced by Dosovitskiy \etal in \cite{Dosovitskiy_Chairs2015} for generating 2D images from 3D objects under different viewpoint angles and various basic transformations. The same architecture was later extended to another application involving face generation\footnote{https://github.com/zo7/deconvfaces} by Michael D. Flynn. In this work, we use the same approach (and architecture) to GNNs %generative neural networks 
and train our own network for deidentification. 

As already suggested in the previous section, we want the network to be able to generate surrogate faces of identities or mixtures of identities contained in the feature database of our deidentification pipeline. Thus, we use RaFDB for GNN training. Once fully trained, the network is able to generate new artificial faces in accordance with the supplied identities,
%identity information (i.e., one or more identities from the training dataset), 
but also in line with other appearance-related parameters that are exposed during the training stage. The generation process can be described as follows:
\begin{equation}
%\mathbf{x} = \text{G-CNN}(\mathbf{y},\mathbf{z}),
\mathbf{x} = \text{GNN}(\mathbf{y},\mathbf{z}),
\end{equation}
where $\mathbf{x}$ is the output of the generative network, GNN, $\mathbf{y}$ stands for an identity-related parameter vector that encodes information about the $k$-closest identities returned by our matching procedure, and $\mathbf{z}$ denotes a parameter vector that guides the generation process and affects specific characteristics of the visual appearance of the generated output. 

In the above equation, the vector $\mathbf{z}$ can in general relate to any appearance characteristic that is appropriately annotated in the training data. Since RaFDB is annotated with respect to different facial expressions, 
we train our network to generate faces with different identities as well as facial expressions. However, the number of appearance-related parameters exposed by the network is not limited and is in general defined by the labels available with the training data.

The generative network consists of fully-connected and deconvolutional layers as described in detail in~\cite{Dosovitskiy_Chairs2015}. Each deconvolution layer includes one upscaling layer followed by a convolution layer. Stabilization of the error during training is achieved by adding batch normalization layers and leaky-rectifier-unit-activation functions between the deconvolution layers. The loss of the training procedure is calculated as the pixel-wise mean square difference between the ground truth image and the artificially generated image. The training is performed with 456 images from the RaFDB dataset on a desktop PC with Intel(R) Core(TM) i7-5820K CPU (3.30GHz), 32GB of RAM, utilizing a TitanX GPU and takes around 24 hours. %The network was trained with backpropagation using stohastic gradient descent with a batch size of 16, 500 epochs, a learning rate of 0.001, and with the Adam optimizer.

\begin{figure}[t]
\centering
\includegraphics[width=1\textwidth]{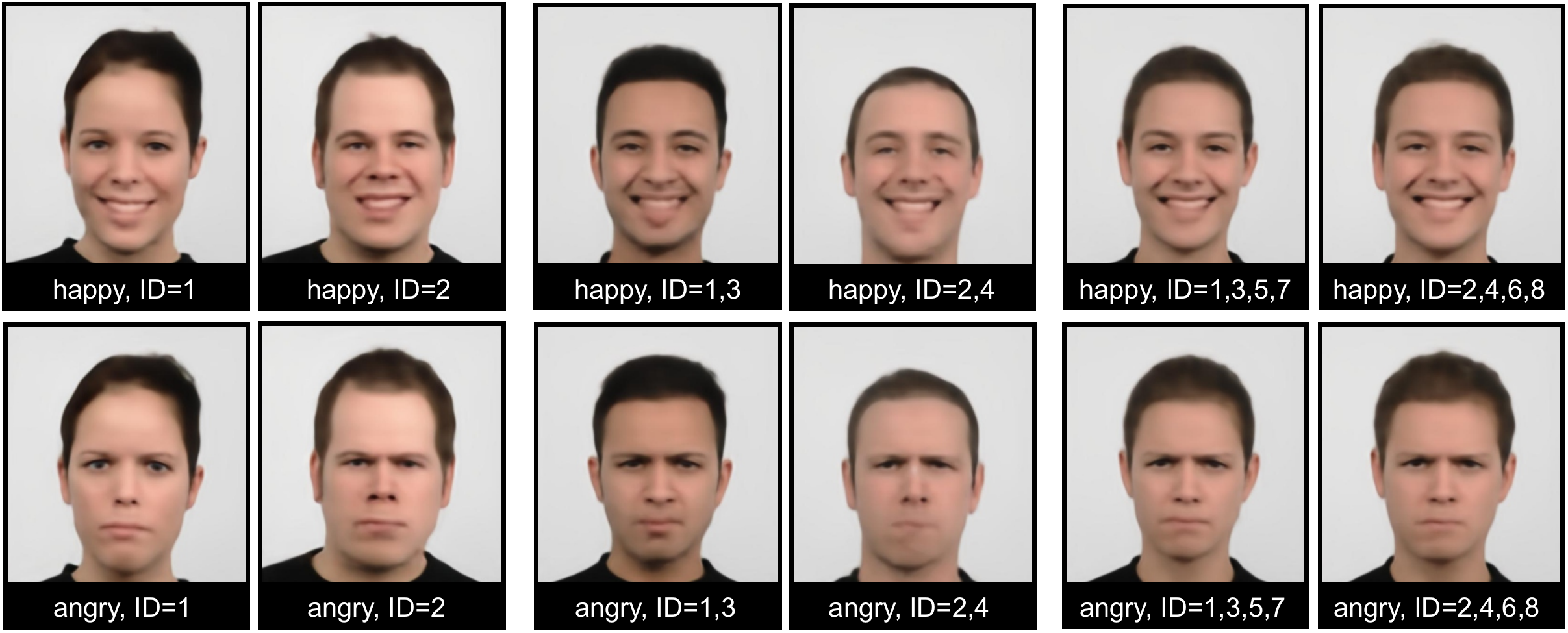}
\caption{Sample outputs from the generative network -- identity mixing with $k$ identities using $k=1$, $k=2$ and $k=4$ is displayed from left to the right, respectively. With an increasing number of identities the generated faces converge towards an average appearance, yet they are still realistic and without ghosting effects. The first label below each images refers to the generated facial expression and the second to the identity / identities used during the generation process.}
\label{fig:generated_outputs}
\end{figure}

With the current training dataset, our network is able to interpolate between different identities as well as different facial expressions. By combining identities, it can generate new averaged faces, almost without ghosting effect as shown in Fig.~~\ref{fig:generated_outputs}. By using the appearance-related parameter vector, $\mathbf{z}$, it can also preserve the utility of the input data, e.g., in the form of facial expressions, which can be retained or altered in regards to the original input. A few sample faces, generated with our GNN are shown in Fig.~\ref{fig:generated_outputs}. Here, the first block of images shows artificial images generated based on a single identity (i.e., $k=1$), the second block shows images computed based on two identities (i.e., $k=2$) and the last block shows sample faces generated from four identities (i.e., $k=4$). Note that the generated faces get closer to an average appearance as the number of identities increases, yet they still appear realistic and feature no ghosting effects.
%The data used to train the network was obtained from the Radbound Faces Database \cite{Langner_RAFDB2010}. The current input parameters to the network are the identity (in form of one-hot encoding vector which denotes involved identities) and emotion (in form of a predefined vector weights, although a different combinations of these settings are also possible).

%anonymity comes from

\subsection{Face replacement}

The last step of our deidentification pipeline is face replacement, during which the generated surrogate face is blended into the input image. The face replacement step starts with facial-landmark estimation using the approach from \cite{Kazemi_One2014}. The landmarks are detected in both, the generated face and the detected original, input face. Using both sets of landmarks, we then estimate a perspective transformation that aligns the landmarks of the artificially generated face and the landmarks of the original face using RANSAC. The generated face image is then warped using this transformation in order to adjust the synthetic content over the landmarks of the original image. This correction is needed in all cases, where faces in the input images are not entirely frontal.

\begin{figure}[tb]
\centering
\includegraphics[width=1\textwidth]{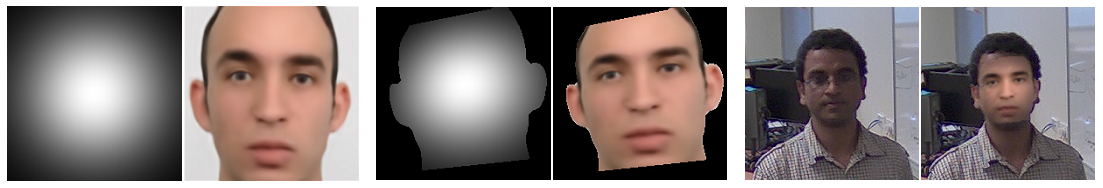}
\caption{Illustration of the replacement procedure (from left to right): the Gaussian mask, the artificially generated face image, the modified Gaussian mask, geometrically corrected synthetic face without background, sample frame, and deidentified frame.}
\label{fig:rep}
\end{figure}

Following the geometric corrections, we apply a second post-processing procedure that discards the background of the generated faces. During this step, simple skin-color segmentation is performed using the \textit{upper} and \textit{lower} boundaries in the HSV color-space that define the skin intensities, i.e., \textit{lower}=$[0, 10, 20]$, \textit{upper}=$[200, 255, 255]$. Pixels with values within the defined range are retained and the rest is discarded. Erosion and dilation are then used to remove possible isolated regions that do not belong to the facial area. With this step we make sure that most of the background around the generated facial area is removed and only the facial region without the gray-colored background, as seen in Fig. \ref{fig:rep}, is swapped during deidentification. %While this procedure works reasonably well, it would also be possible to tweak the generative part of the pipeline to produce facial segmentation masks. However, this would require additional architectural changes as well as accurately annotated (at the pixel level) (ground truth) segmentation masks. 

In the last step, the warped and segmented synthetic face image is blended with the original image. Blending is performed with a Gaussian kernel mask: 
\begin{equation}
g(x,y) = e^{{{ - (\left( {x - \mu_x } \right)^2 + \left( {y - \mu_y } \right)^2 ) } \mathord{\left/ {\vphantom {{ - \left( {x - \mu } \right)^2 } {2\sigma ^2 }}} \right. \kern-\nulldelimiterspace} {2\sigma ^2 }}},
\end{equation}
where $ \mu_x = s/2$, $\mu_y = s/2$, $\sigma = s/6$, $s=\min(w, h)$, $w$ and $h$ stand for the dimensions of the generated image, and $x$ and $y$ denote image coordinates. This online generated kernel then serves as a weight mask when blending the original and generated image pixels. The kernel is warped using the same homography transformation in order to ensure the best possible face alignment and a suitable level of naturalness of the final output. The replacement procedure is illustrated in Fig.~\ref{fig:rep}. Here, the first image shows the Gaussian weight mask, the second image shows the initial output of the generative network, the third image shows the Gaussian mask modified with the result of the geometric correction and segmentation step, the fourth image presents the adjusted (synthetic face) and the last two images depict an original and deidentified frame from our test dataset.

%\subsection{Output}

%The output of our pipeline is altered input image. The image is modified in such a way, that every detected face on the original input is replaced with a generated face, that represents a new, artificially generated identity. This new identity is a mixture from a predefined set of face images -- in our case these images are taken from the frontal part of Radbound Faces Database \cite{Langner_RAFDB2010}.

\section{Experiments and results}\label{sec:experimenti}

In this section we present experimental results aimed at demonstrating the merits of our deidentification pipeline. We first discuss the experimental dataset and performance metrics and then present qualitative as well as quantitative results.

\subsection{Dataset, experimental setup and performance measures}\label{sec:dataset}

To evaluate the performance of our deidentification approach, we use the ChokePoint dataset~\cite{Wong_Chokepoint2011}, which contains video footage captured in a typical surveillance scenario. People in the videos were recorded while walking through a portal above which an array of 3 cameras was placed. The ChokePoint videos exhibit variations across illumination conditions, pose, image sharpness, and alike and are well suited for studying the performance of deidentification technology. %A few example frames from the videos in the ChokePoint dataset are shown in Fig.~\ref{fig:CP_samples}.
%\begin{figure}[t]
%\centering
%\includegraphics[width=1\linewidth]{figures/CP_samples.pdf}
%\caption{Example frames from the videos in the ChokePoint dataset~\cite{Wong_Chokepoint2011}. The videos were captured in a typical surveillance scenario and are therefore well suited for deidentification experiments.}
%\label{fig:CP_samples}
%\end{figure}

The ChokePoint dataset contains 48 video sequences with a total of 64,204 frames. The videos feature 25 subjects walking through the first portal and 29 subjects walking through the second portal. For our experiments, we partitioned the video sequences into two distinct subsets. The first subset contained 24 sequences with people in mostly frontal poses, while the second subset contained the remaining 24 sequences with people in less frontal poses, i.e., profile frames. We refer to the former subset as \textit{original} and to the latter as \textit{profile}  from hereon. The video sequences from the original subset were subjected to our deidentification approach and stored for the experimental evaluation.

To measure the efficacy of the developed deidentification pipeline, we conduct four types of verification experiments with a 10-fold cross-validation protocol. During each fold, we  perform 300 legitimate (matching, client) and 300 illegitimate (non-matching, impostor) verification attempts. The different types of verification experiments are briefly outlined below:
\begin{itemize}
\item \textbf{Original vs. original:} In this experiment we sample 300 image pairs for the legitimate verification attempts and 300 image pairs for the illegitimate verification attempts for each experimental fold from video sequences of the original subset. The goal of this experiment is to establish the baseline performance of the recognition techniques considered in our experiments. Since video frames are sampled from the same set of videos, this experiment may be biased towards higher performances, since the appearance variability between frames is limited.   
\item \textbf{Original vs. profile:} In this experiment we construct the image pairs for the legitimate and illegitimate verification attempts of each fold from images taken from the original and profile subsets. Here, the first image in the pair is always sampled from the original subset and the second is always sampled from the profile subset. Because the two subsets contain distinct video sequences, this experiment better reflects the baseline performance of the recognition techniques considered in our experiments.     
\item \textbf{Deidentified vs. original:} This experiment is equivalent to the \textit{original vs. original} experiment with the difference that the first image of each image pair is replaced with its deidentified version. Thus, the experiment is meant to measure the efficacy and performance of the proposed deidentification procedure. All verification attempts of all 10 cross-validation folds in this experiment have a direct correspondence in the original vs. original experiment and, therefore, clearly demonstrate the effect of deidentification on the verification performance. 
\item \textbf{Deidentified vs. profile:} The last experiment follows the same approach as the \textit{original vs. profile} experiment, but replaces the video frames from the original subset with its deidentified version. The goal of this experiment is  again to demonstrate the feasibility of our deidentification approach.   
\end{itemize}

We report performance with standard performance metrics and graphs. Specifically, we present Receiver Operating Characteristic (ROC) curves~\cite{ROC_curves},~\cite{Ziga_Hindawi}, which plot the value of the verification rate (VER) against the false acceptance rate (FAR) for different values of the decision threshold, and report a number of scalar performance metrics for all experiments, i.e., the equal error rate (EER), which is the operating point on a ROC curve, where FAR and 1-VER are equal, the verification rate at 1\% FAR (VER-1) and the area under the ROC curve (AUC)~\cite{Gajsek_ROC}. Because we use a 10-fold cross validation protocol, we report all metrics in the form of the mean and standard deviation computed over all experimental folds~\cite{Neurocomputing}.   

\subsection{Qualitative evaluation}

\begin{figure}[tb]
\centering
\begin{minipage}{\textwidth}
  \centering
  \includegraphics[width=.163\linewidth]{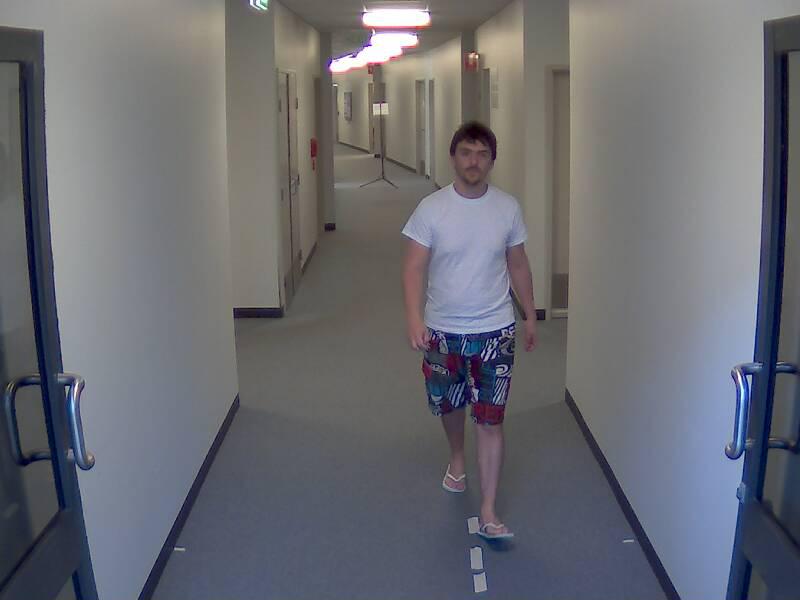}
  \hspace{-0.2cm}
  \includegraphics[width=.163\linewidth]{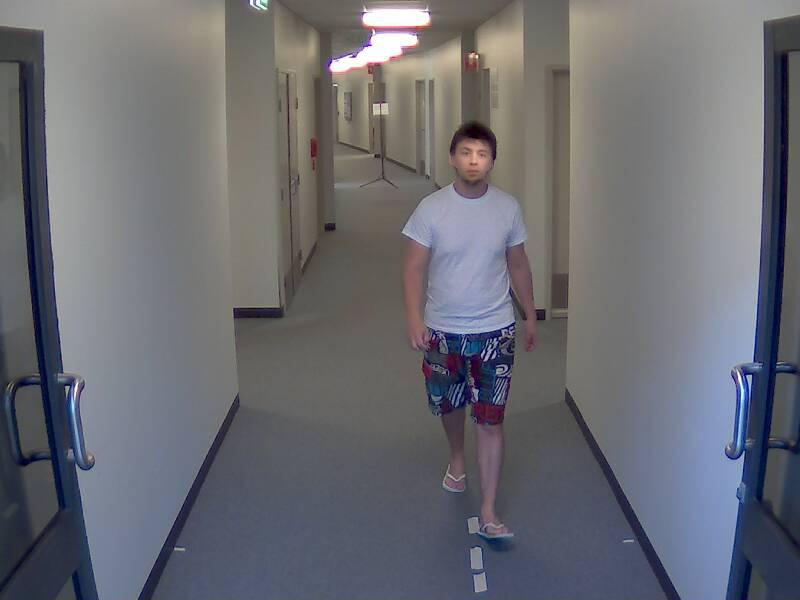}
  \includegraphics[width=.163\linewidth]{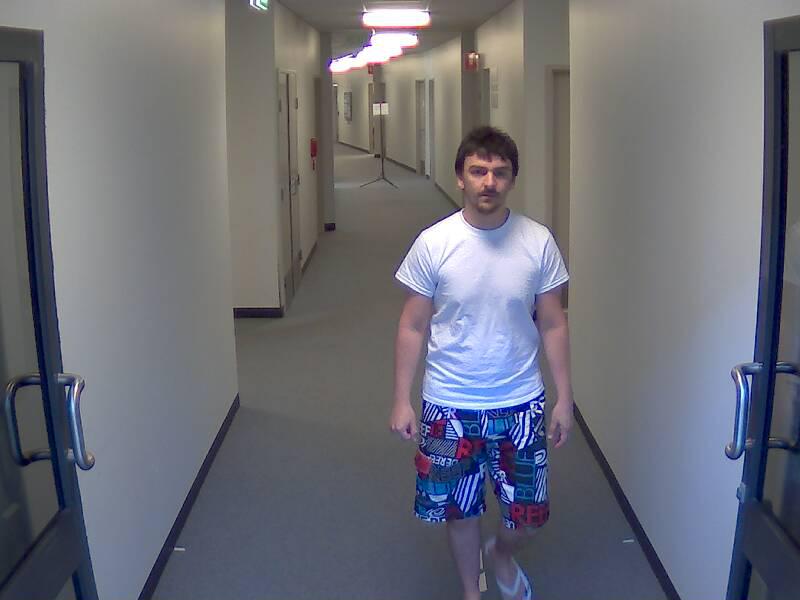}
  \hspace{-0.2cm}
  \includegraphics[width=.163\linewidth]{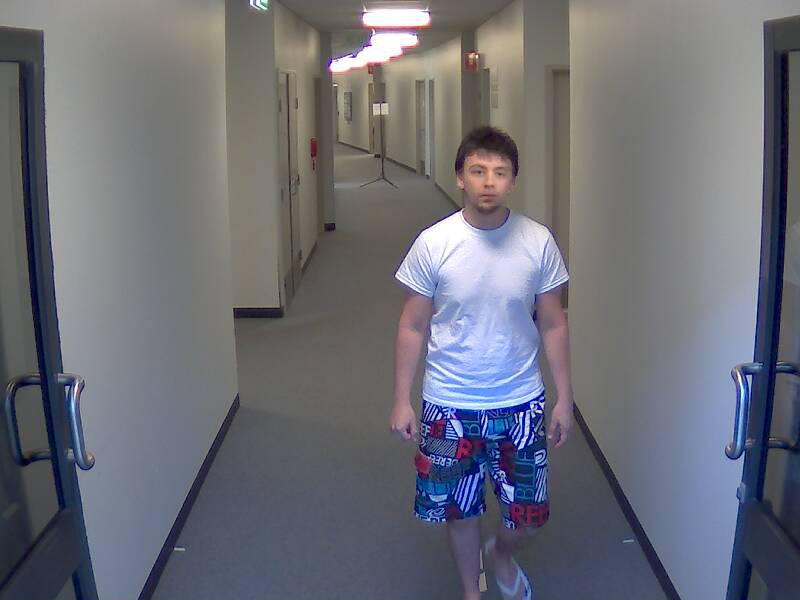}
  \includegraphics[width=.163\linewidth]{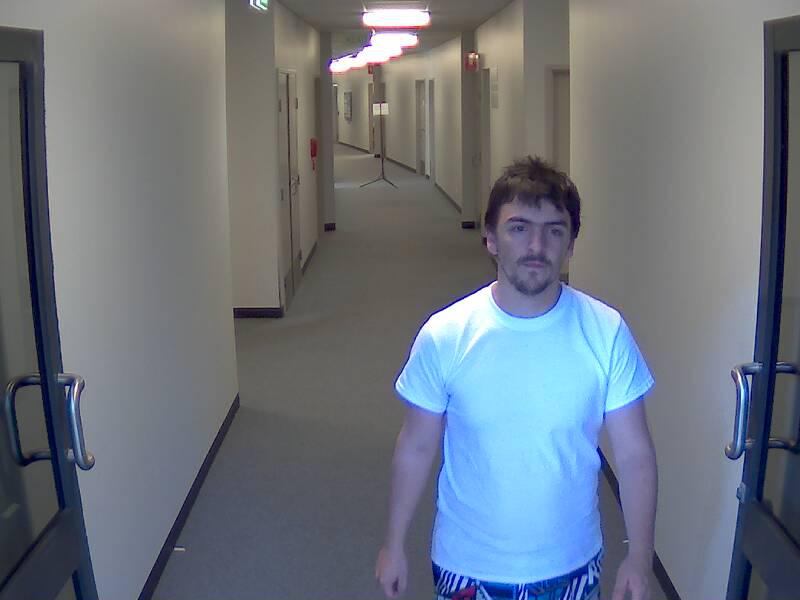}
  \hspace{-0.2cm}
  \includegraphics[width=.163\linewidth]{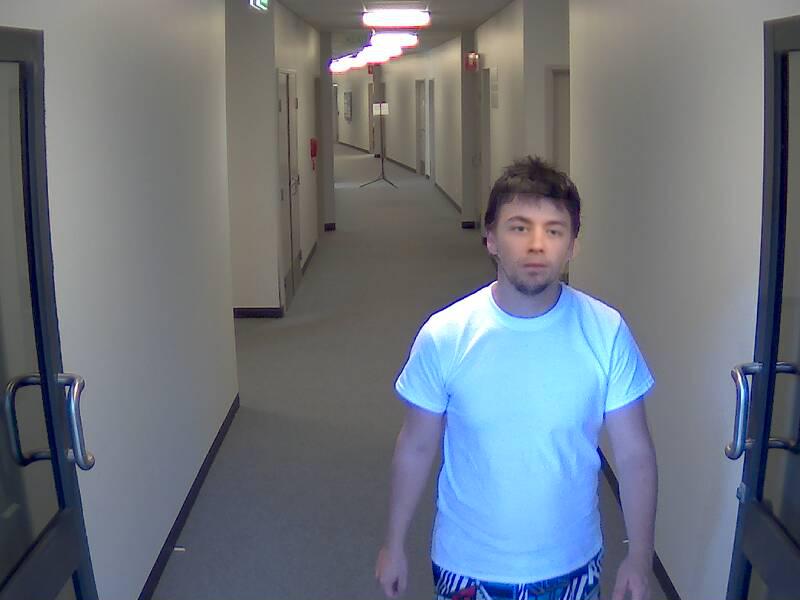}
\end{minipage}%
\vspace{0.05cm} \\
\begin{minipage}{\textwidth}
  \centering
  \includegraphics[width=.163\linewidth]{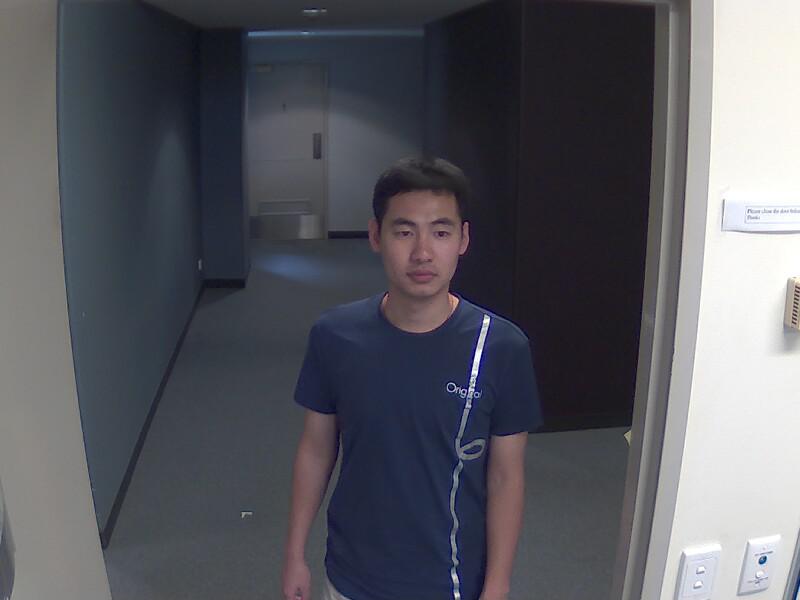}
  \hspace{-0.2cm}
  \includegraphics[width=.163\linewidth]{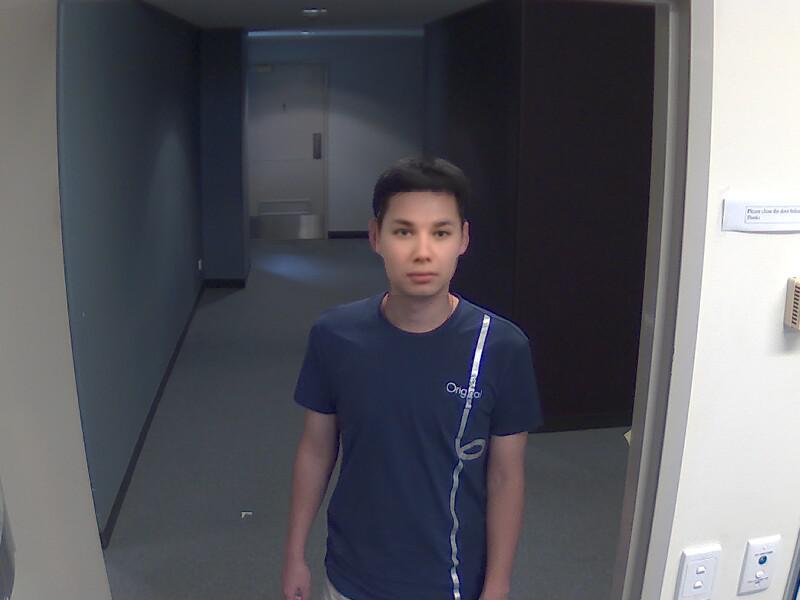}
  \includegraphics[width=.163\linewidth]{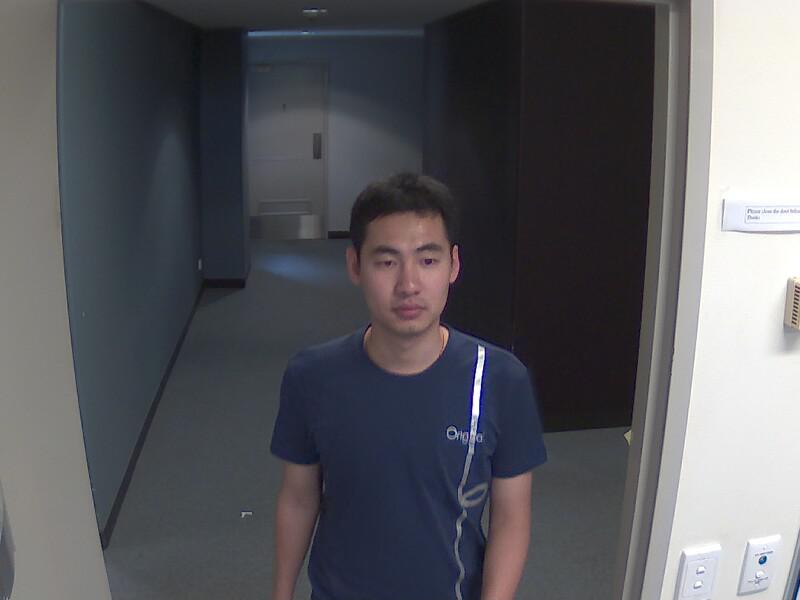}
  \hspace{-0.2cm}
  \includegraphics[width=.163\linewidth]{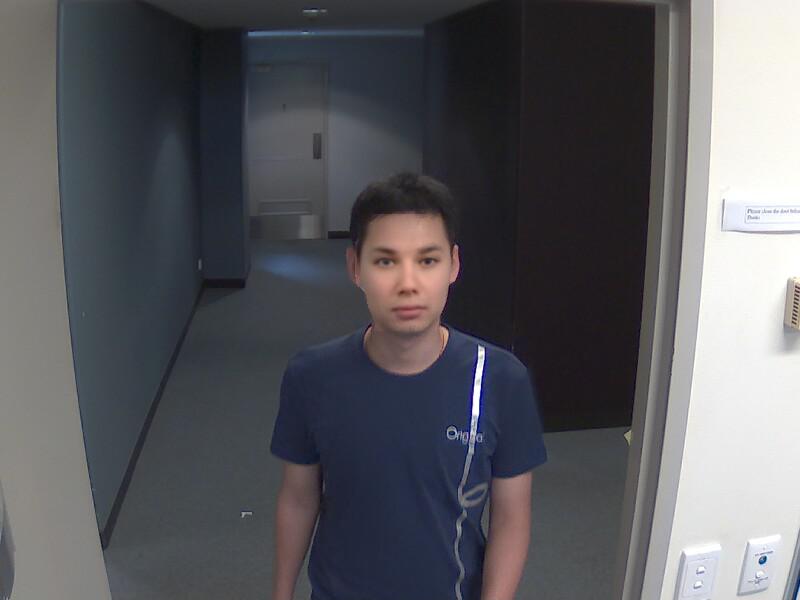}
  \includegraphics[width=.163\linewidth]{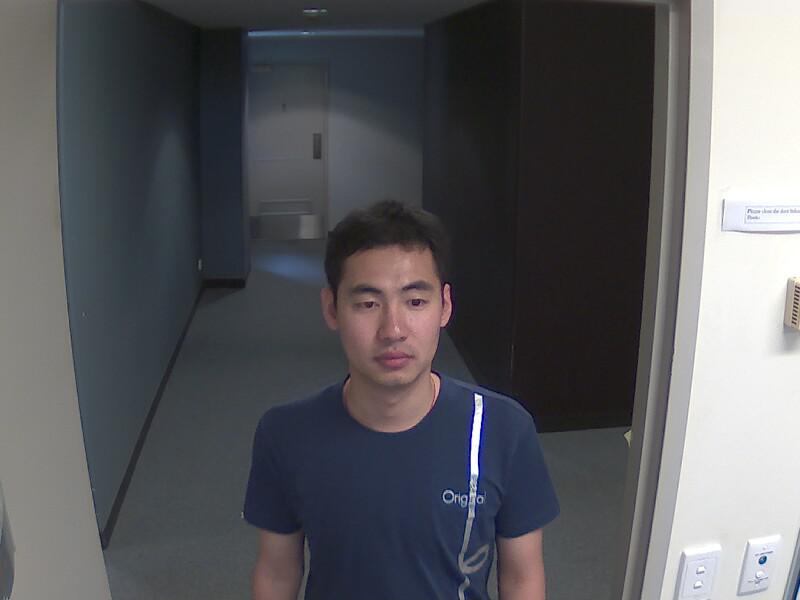}
  \hspace{-0.2cm}
  \includegraphics[width=.163\linewidth]{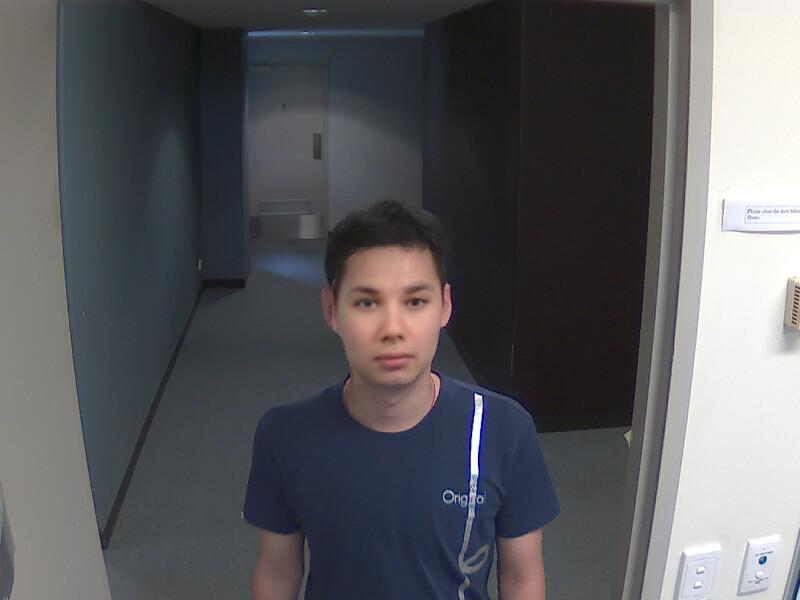}
\end{minipage}%
\caption{Qualitative examples of deidentified frames. Each row shows a few example frames from a video sequence of the ChokePoint dataset (left image of each pair) and the corresponding deidentification results (right image of each pair). Note how the generative network is able to generate realistic renderings of faces for deidentification.}
\label{fig:deid_examples_good}
\end{figure}

We first demonstrate the efficacy of our deidentification approach with a few qualitative examples in Fig.~\ref{fig:deid_examples_good}. Here, each row shows a few frames from a video sequence of the ChokePoint dataset and the corresponding deidentification result. In each image pair, the left image represents the original frame and the right image its deidentified counterpart. We can see that for the most part the deidenfied faces generated by the generative network appear natural and realistic. %\textcolor{red}{
In this case, substituted identities are generated from 2 most similar identities (i.e.  $k=2$). 
%}
% R1

\begin{figure}[tb]
\centering
\includegraphics[width=1\textwidth]{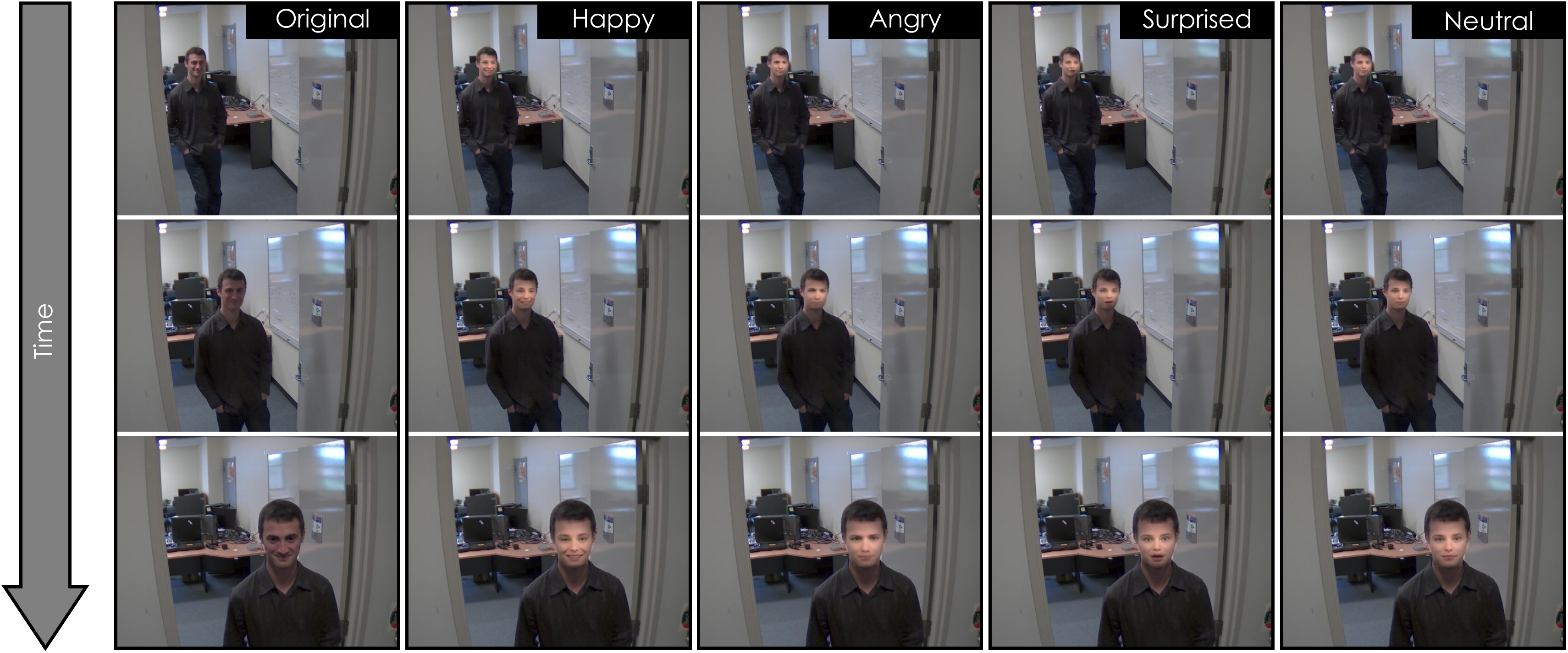}
%\begin{minipage}{\textwidth}
%  \centering
%   \includegraphics[width=.24\linewidth]{orig21}
 %  \includegraphics[width=.24\linewidth]{deid-happy-1}
  % \includegraphics[width=.24\linewidth]{deid-angry-1}
  % \includegraphics[width=.24\linewidth]{deid-surprised-1}
% \end{minipage}%
% \vspace{0.05cm} \\
% \begin{minipage}{\textwidth}
 %  \centering
  % \includegraphics[width=.24\linewidth]{orig22}
  % \includegraphics[width=.24\linewidth]{deid-happy-2}
  % \includegraphics[width=.24\linewidth]{deid-angry-2}
  % \includegraphics[width=.24\linewidth]{deid-surprised-2}
% \end{minipage}
% \vspace{0.05cm} \\
% \begin{minipage}{\textwidth}
 %  \centering
  % \includegraphics[width=.24\linewidth]{orig23}
  % \includegraphics[width=.24\linewidth]{deid-happy-3}
 %  \includegraphics[width=.24\linewidth]{deid-angry-3}
 %  \includegraphics[width=.24\linewidth]{deid-surprised-3}
% \end{minipage}
\caption{Deidentified frames rendered with different facial expressions. Images in the first column represent original frames from a video sequences of the ChokePoint dataset. The second, third, fourth and fifth columns show deidentified frames rendered with a ``happy'', ``angry'', ``surprised'' and ``neutral'' expression, respectively. As can be seen, our deidentification approach is highly flexible and is able to retain or alter specific aspects of the deidentified faces.}
\label{fig:deid_examples_happy}
\end{figure}

One key characteristic of our deidentification approach is the flexibility that the generative network offers when producing synthetic face images for deidentification. The generation process can be parameterized with respect to the desired target appearance of the synthetic face, which makes it possible to generate  faces with different characteristics (in terms of facial expression, skin color, gender, etc.) and is important when trying to retain non-identity-related information in the deidentified  data. In video conferencing applications, for example, one may want to protect the privacy of the conference participants by hiding their identity, but still preserve the information that facial expressions convey during the conversation. In customer-profiling applications, the focus is typically on the demographics of the customers (such as gender or age distributions) and not on the identity. With our deidentification approach we are able to conceal the identity of people in the image data and retain (or alter) certain aspects of the facial appearance. This characteristic is demonstrated in Fig.~\ref{fig:deid_examples_happy}. Here, the first column shows a few sample frames from a video sequence of the ChokePont dataset and the second, third, fourth and fifth column show three different deidentification results that were rendered with a ``happy'', ``angry'', ``surprised'' and ``neutral'' facial expression, respectively. While we only show  results for different facial expressions, our deidentification pipeline is in general able to generate variations of synthetic faces in accordance with any appearance-related label of the training data. Thus, if the data used for training contains images annotated with respect to facial expressions, we are able to generate faces with different facial expressions, if the data contains labels for gender, we can synthesize faces belonging to males or females and so forth. The number of different appearance variations our approach can cover is only limited by the number of available labels. 

%Thus, our approach is able to generate synthetic faces for deidentification with different emotions, For example, while we ma %napiši kaj o aplikacijah, ki skenirajo kupce v shoping centrih in jih zanima starost, spol, rasa, ipd ne pa tudi identiteta.   , video conferencing  - convey information pertaining to emocije ipd..... 

\begin{figure}[tb]
\centering
\begin{minipage}{\textwidth}
  \centering
  \includegraphics[width=.164\linewidth]{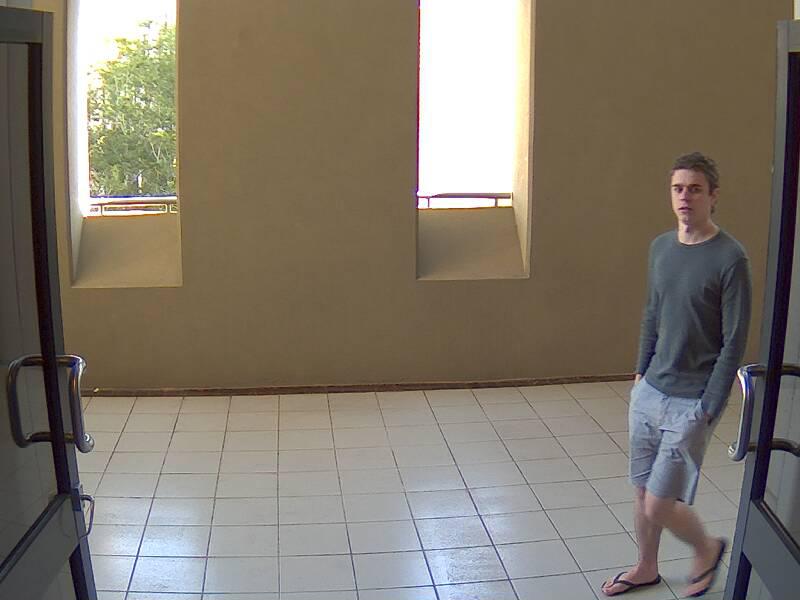}
  \hspace{-0.2cm}
  \includegraphics[width=.164\linewidth]{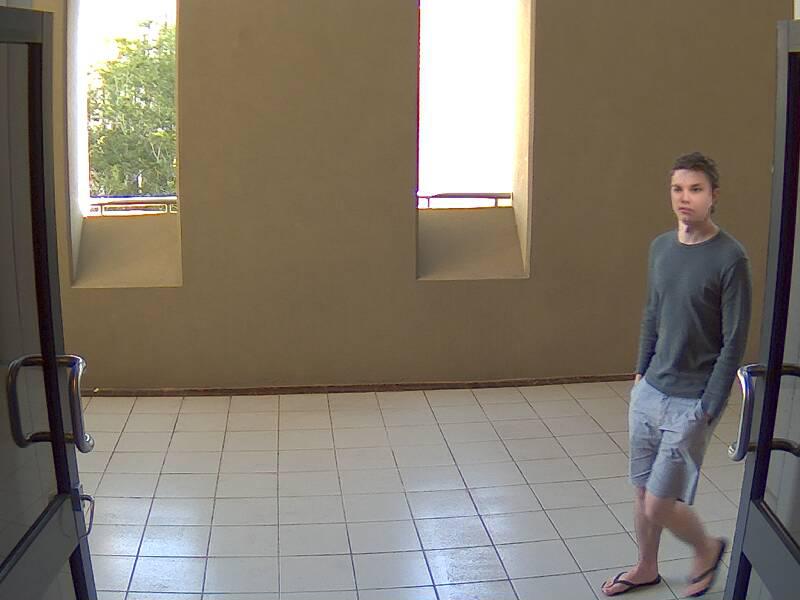}
  \includegraphics[width=.164\linewidth]{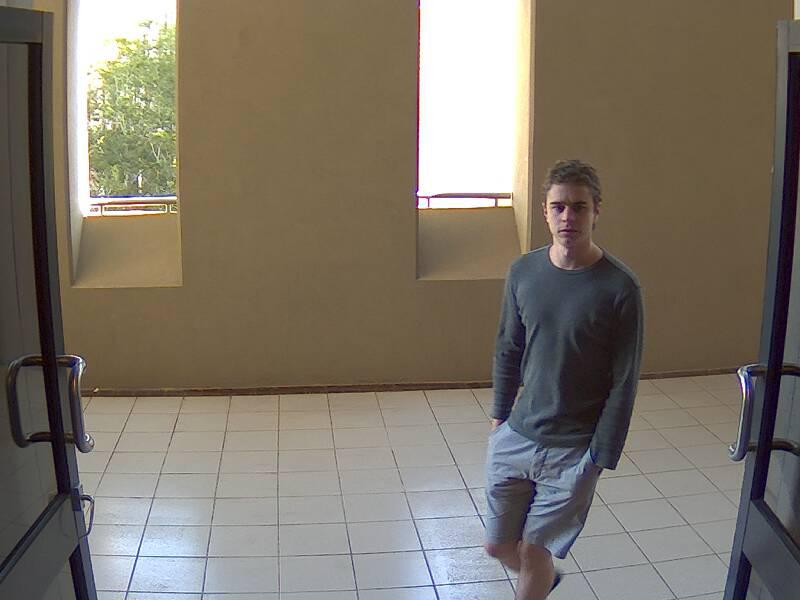}
  \hspace{-0.2cm}
  \includegraphics[width=.164\linewidth]{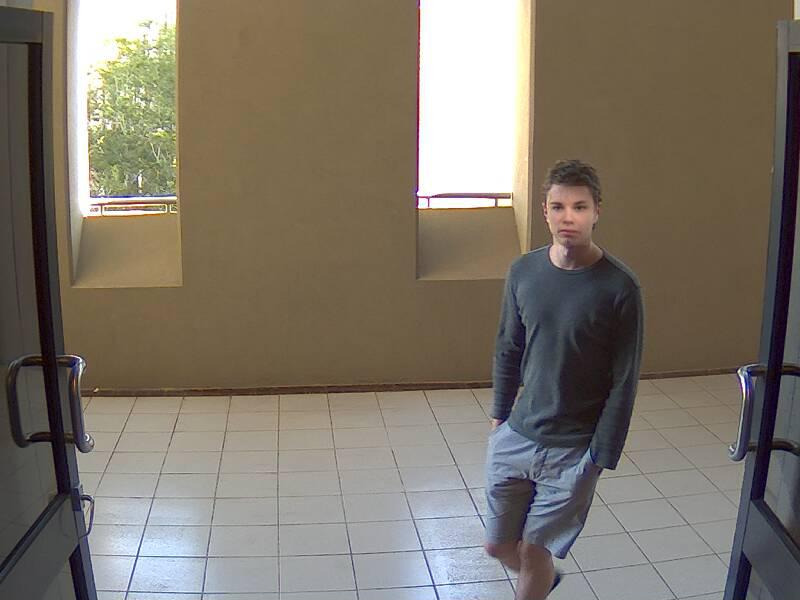}
  \includegraphics[width=.164\linewidth]{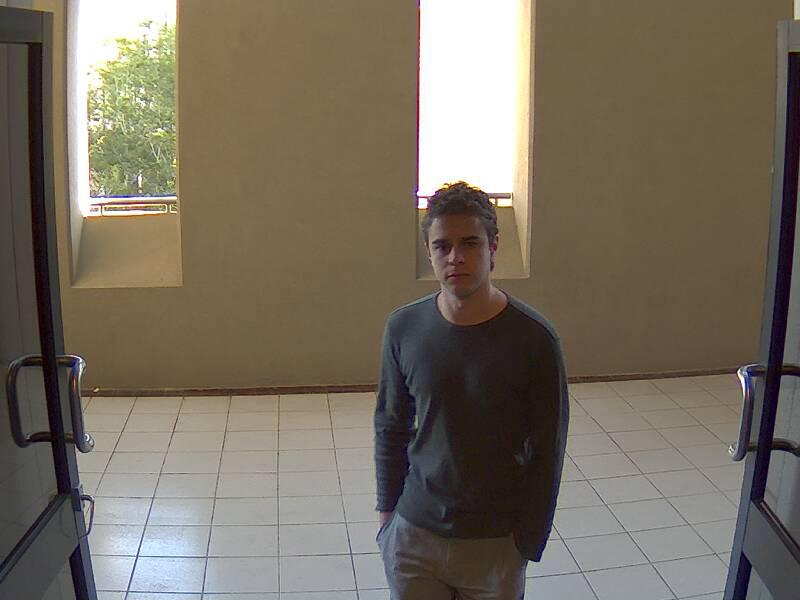}
  \hspace{-0.2cm}
  \includegraphics[width=.164\linewidth]{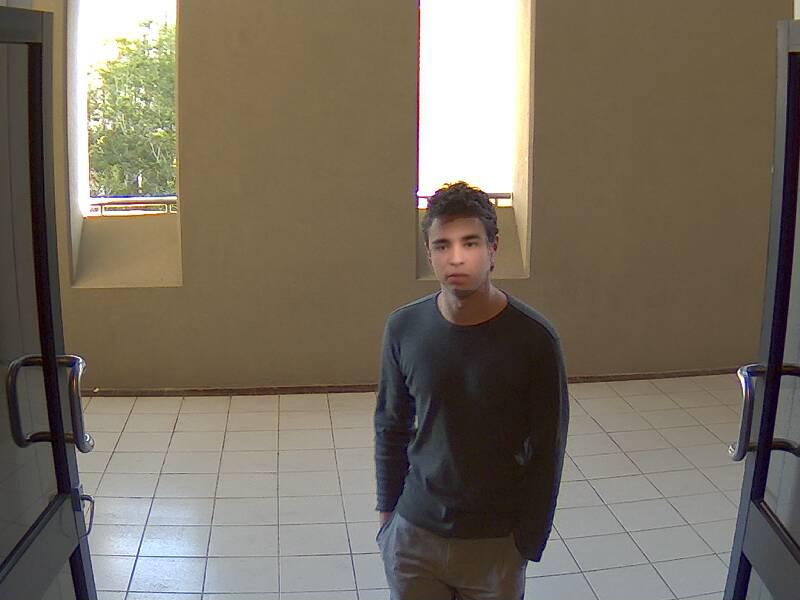}
\end{minipage}
\vspace{0.05cm} \\
\begin{minipage}{\textwidth}
  \centering
  \includegraphics[width=.164\linewidth]{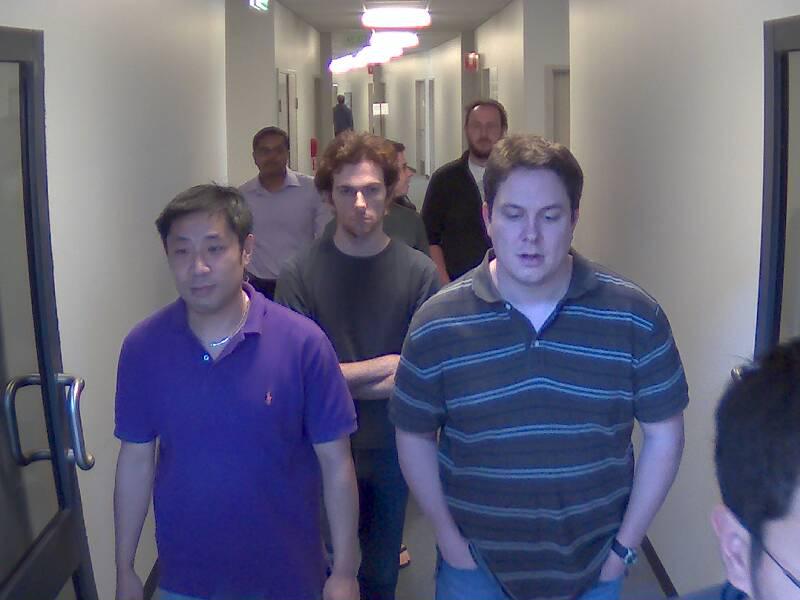}
  \hspace{-0.2cm}
  \includegraphics[width=.164\linewidth]{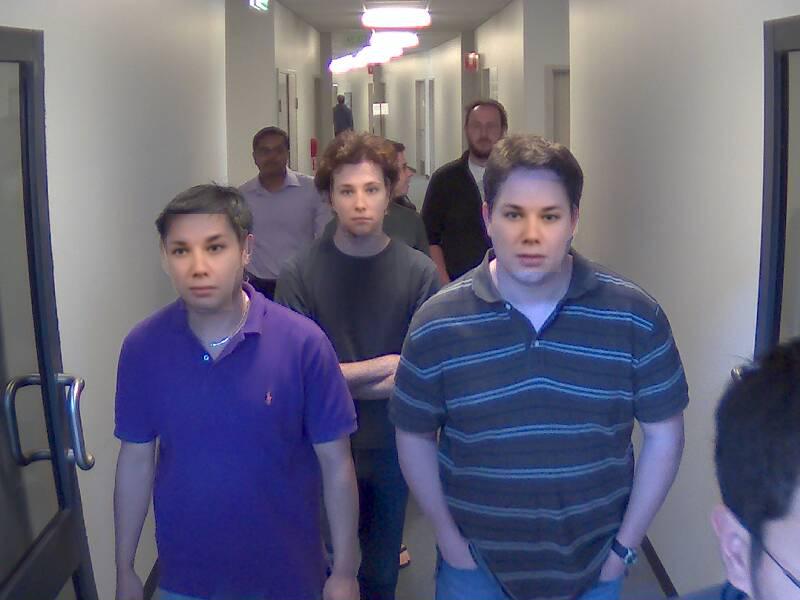}
  \includegraphics[width=.164\linewidth]{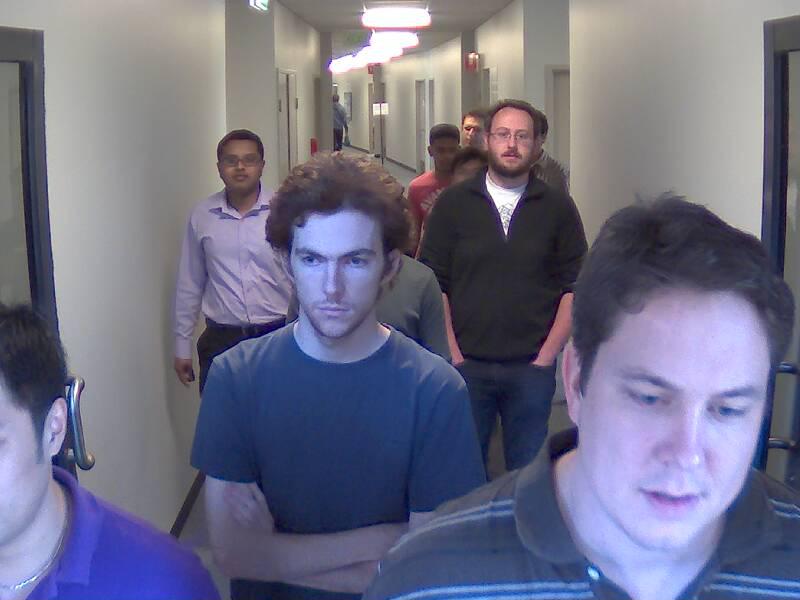}
  \hspace{-0.2cm}
  \includegraphics[width=.164\linewidth]{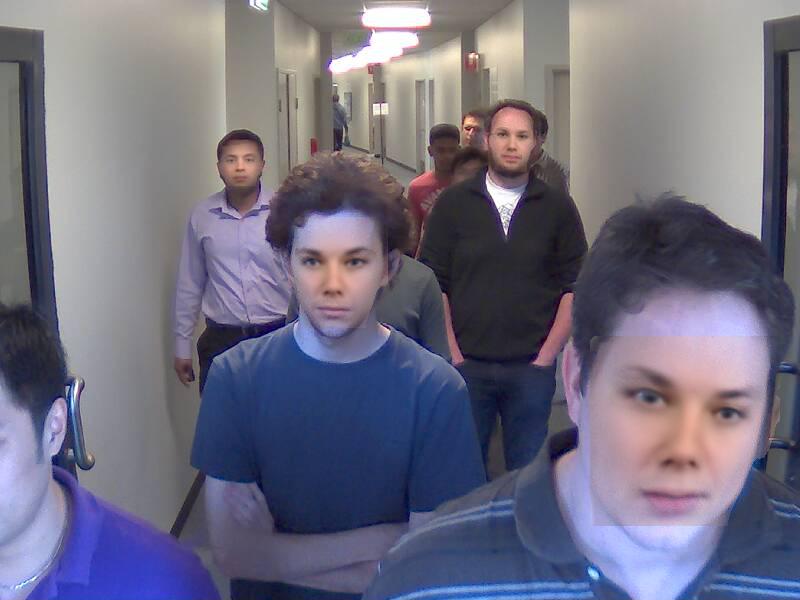}
  \includegraphics[width=.164\linewidth]{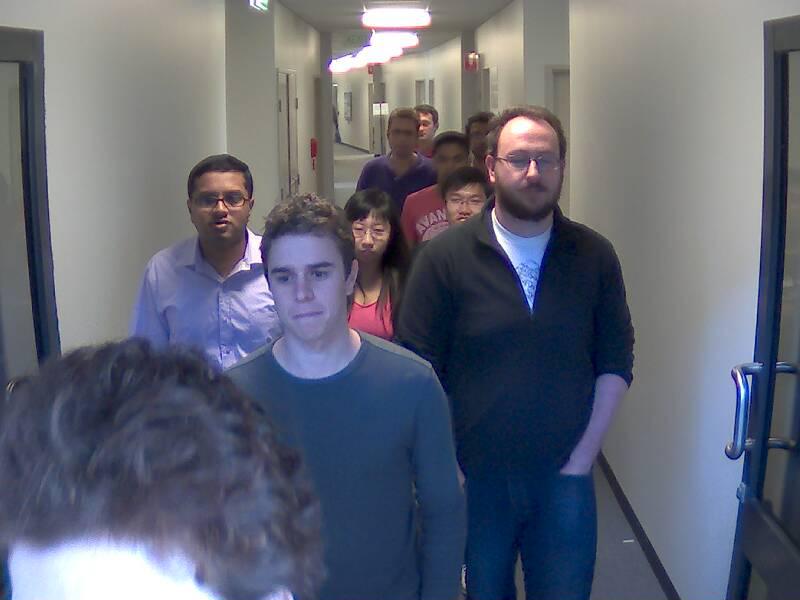}
  \hspace{-0.2cm}
  \includegraphics[width=.164\linewidth]{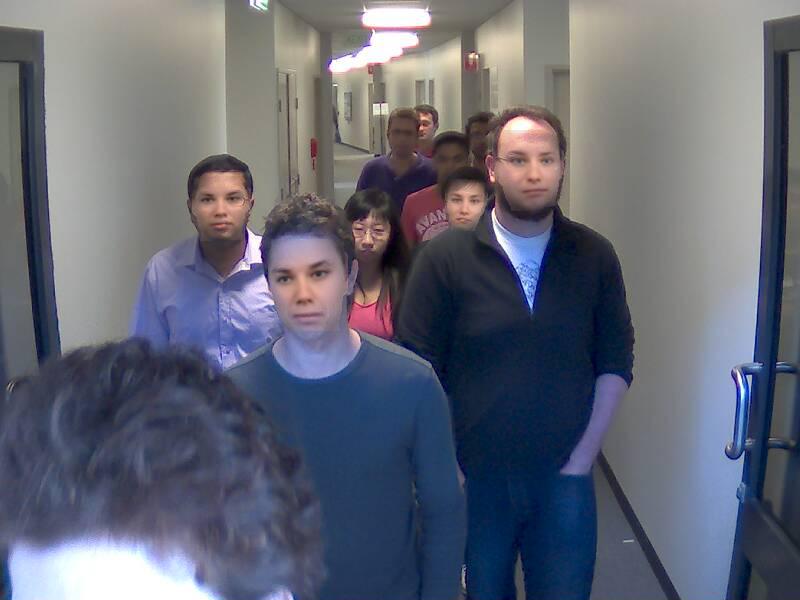}
\end{minipage}
\caption{Qualitative examples of problematic deidentification results. The upper row shows an identity switch in the last frame due to a change in the scenes illumination. The lower row shows difficulties due to the presence of multiple people, some of which occlude faces in the background. Misalignment between the original and surrogate faces is also visible in the second image pair of the lower row, which happens due to extreme viewing angles when people exit the scene.}
\label{fig:deid_examples_bader}
\end{figure}

In Fig.~\ref{fig:deid_examples_bader} we show some examples of visually less pleasing (or problematic) deidentification results. The image artifacts visible here are a consequence of different scene conditions (see the fourth image in the second row of Fig.~\ref{fig:deid_examples_bader} for an extreme example) and can be ascribed to our replacement procedure. These artifacts could be alleviated by a more elaborate face-replacement approach exploiting, for example, Poisson blending or color-profile matching. However, this would affect the speed of our pipeline, which currently runs at around 12 frames per second if processing the sequence with only one subject present at the time and around 5 frames per second if executing it on a sequence involving multiple subjects simultaneously present in a scene. These framerates were achieved on a desktop PC with Intel(R) Core(TM) i7-6700K CPU (4.00GHz) and 32GB of RAM. Another cause of image artifacts are extreme facial poses when people exit the scene (this is common to all sequences), which result in visible misalignment between the superimposed (deidentified) faces and the original facial areas.    

Among the main limitations of our deidentification approach is the persistence of identity in the deidentified data. As we are dealing with video footage, the deidentification procedure should ideally produce the same (consistent) result for all frames of the given video sequence. In other words, the facial area of a given subject should be replaced with an artificially generated face of the same target identity over the entire duration of each video. However, due to changes in facial appearance, our matching module occasionally returns inconsistent results and causes changes in the target identity of the deidentified frames. This effect is demonstrated in the first row of Fig.~\ref{fig:deid_examples_bader}, where an identity change can be observed in the last image pair due to variations in the scene's illumination despite the fact that the same subject is being deidentified. Nevertheless, because the target identity for deidentification is determined with a state-of-the-art face recognition model (i.e., VGG~\cite{Parkhi_Recog2015}), our procedure is able to assign a consistent target identity most of the time for all test videos considered in our experiments. % we see that the scene's illumination changes during the course of the video sequence. This change affects the result of our matching module and consequently causes 
%can see intensive change of scene illumination, which in this case caused identity shift, because matching component picked the most similar identity based on the current input. 

%Minor difficulties also appear with sequences of multiple subjects (shown in the second row of Fig.~\ref{fig:deid_examples_bader}), where replacing the facial area of people in the foreground sometimes occludes faces of people in the background. In these sequences partially occluded face may also not be deidentified until they are fully visible due to limitations of the face detector. While this may not be problematic for automatic applications, which also rely on a face detector to perform recognition, such faces may be potentially identified by human operators screening the video footage. 

In Section \ref{sec:con}, where we discuss possible directions for future work, we propose  some possible improvements of our deidentification pipeline, which address most of the existing issues of the current implementation.

%\begin{figure}[t]
%\centering
%\includegraphics[width=1\linewidth]{figures/deid_examples.pdf}
%\caption{Qualitative examples of deidentified frames. Each row shows a few example frames from a video sequence of the ChokePoint dataset (left image of each pair) and the corresponding deidentification result (right image of each pair). Note how the generative network is able to generate realistic renderings of faces for deidentification.}
%\label{fig:deid_examples}
%\end{figure}

%\begin{figure}[t]
%\centering
%\includegraphics[width=1\linewidth]{figures/de_id_examples_happy.pdf}
%\caption{Qualitative examples of deidentified frames rendered with a non-neutral emotional state. Again the left image in each pair represents the original image and the right image shows the deidentified version. The first row shows the same frames as  the last row of Fig.~\ref{fig:deid_examples}, but exhibiting a ``happy'' emotion. The second row shows a few examples with  ... other emotions??.}
%\label{fig:deid_examples_happy}
%\end{figure}

\subsection{Automatic and manual reidentification}

The quality and efficacy of deidentification techniques is typically measured through reidentification experiments~\cite{NIST_Simson}, where the goal is to evaluate the risk of successfully identifying a person from deidentified data. This risk is commonly assessed with automatic and manual recognition experiments. 

As outlined in Section~\ref{sec:dataset}, we perform a number of verification experiments in a 10-fold cross validation protocol towards the risk assessment and consider three state-of-the-art automatic recognition approaches from the literature. Specifically, we use \textit{i)} the open-source implementation of the 16-layer VGG face network from~\cite{Parkhi_Recog2015} 
%as the first recognition approach  
-- VGG from hereon, \textit{ii)} our own 24-layer implementation of the SqueezeNet network from~\cite{SqueezeNet} trained on around 2.5 million images (i.e., the VGG network training data) 
-- SqueezeNet from hereon, 
%-- as the second recognition approach, 
and \textit{iii)} the 4SF algorithm from OpenBR (version 1.1) \cite{Klontz2013_OpenBR}
%as the third recognition approach 
-- OpenBR from hereon. For the two networks, we use the output from the last fully connected layer of each network as a feature vector and compute a similarity score for an image pair as the cosine angle between the two corresponding feature vectors. For OpenBR we use the default matching option. 

We also conduct manual recognition experiments using a similar 10-fold cross-validation protocol as with automatic techniques, but limit the extend of comparisons to 5\% of the automatic experiments. Thus, 30 comparisons are performed during each fold resulting in a total of 300 verification attempts (150 legitimate and 150 illegitimate experiments) in each experimental run. Each verification experiment was evaluated by one human evaluator, i.e., four evaluators covered four verification experiments. To produce similarity scores needed for generating performance metrics and ROC curves, we manually assign a similarity score from a five-point scale to each comparison in accordance with the methodology proposed in~\cite{HumanAnot}. 

Similar to other existing works on face deidentification, 
% dej se vec reference od Gros-a, Sweeneyeve, itd. 
our approach tries to conceal the identity of people by replacing the detected facial areas with a synthetically generated surrogate faces. However, identity cues can also be extracted from contextual information that is not directly related to facial appearance. For example, the facial outline, hair-style, or even clothing can represent a give-away that recent recognition techniques based on deep models as well as humans may be able to pick up. To explore this issue, we conduct two sets of experiments: 
\begin{itemize}
\item With context: here we feed the facial area to the recognition technique directly as it is detected by the face detector. Thus, the facial area also contains contextual information about the shape of the head, hair style and alike. A comparison of two images with context is illustrated in the last column of Table~\ref{Tab:_numeric results} (first row). 
\item Without context: here we trim the bounding box returned by the face detector on each side by 10\%, the facial areas used for the recognition experiments are therefore cropped tighter and contain only little contextual information. A sample comparison of two images as used in this set of experiments is shown in the last column of Table~\ref{Tab:_numeric results} (second row).   
\end{itemize}

\begin{figure*}[!thb]
\begin{minipage}{0.5\textwidth}
 \centering
 \includegraphics[width=1\textwidth]{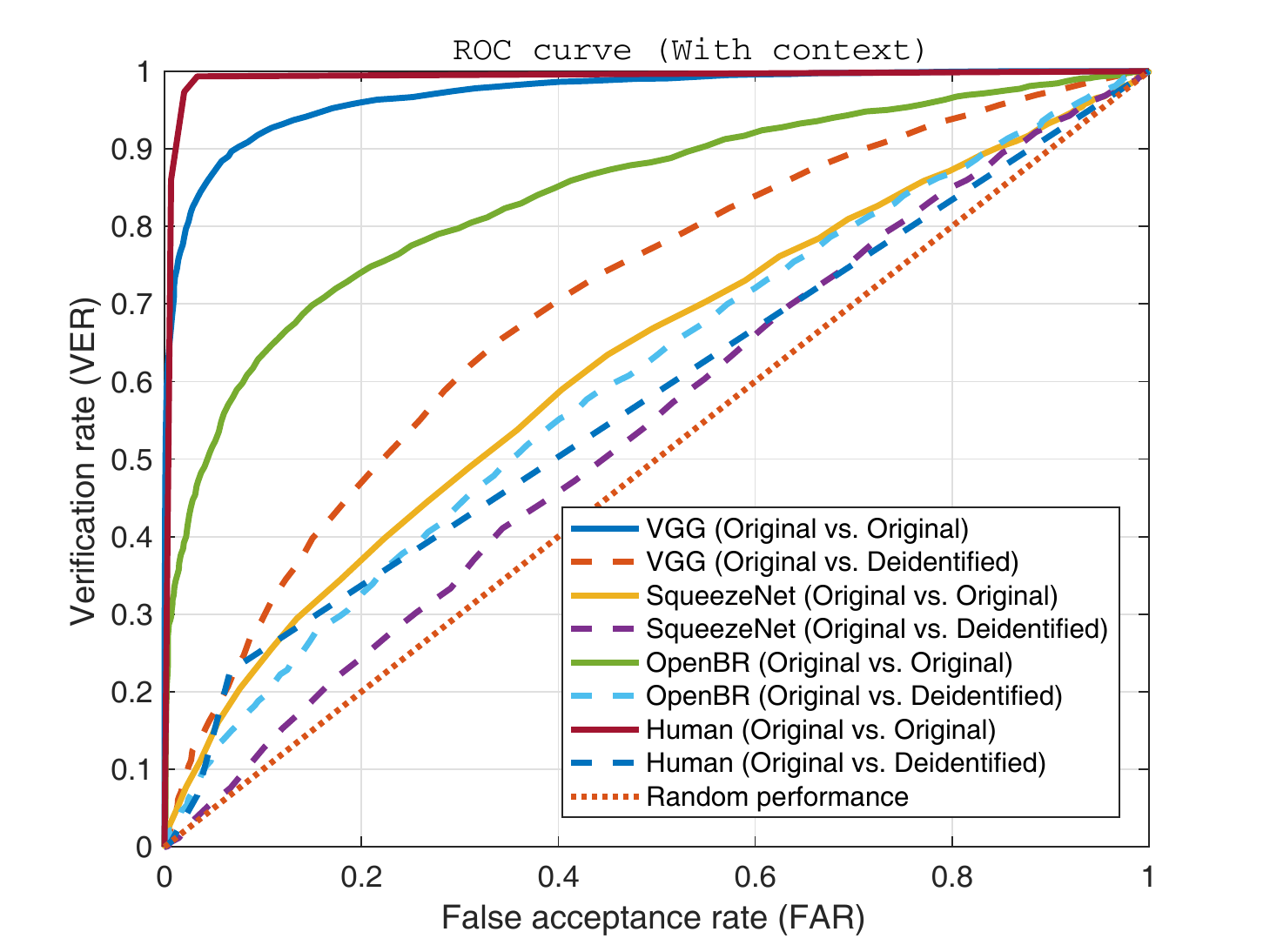}
\end{minipage}
\begin{minipage}{0.5\textwidth}
 \centering
 \includegraphics[width=1\textwidth]{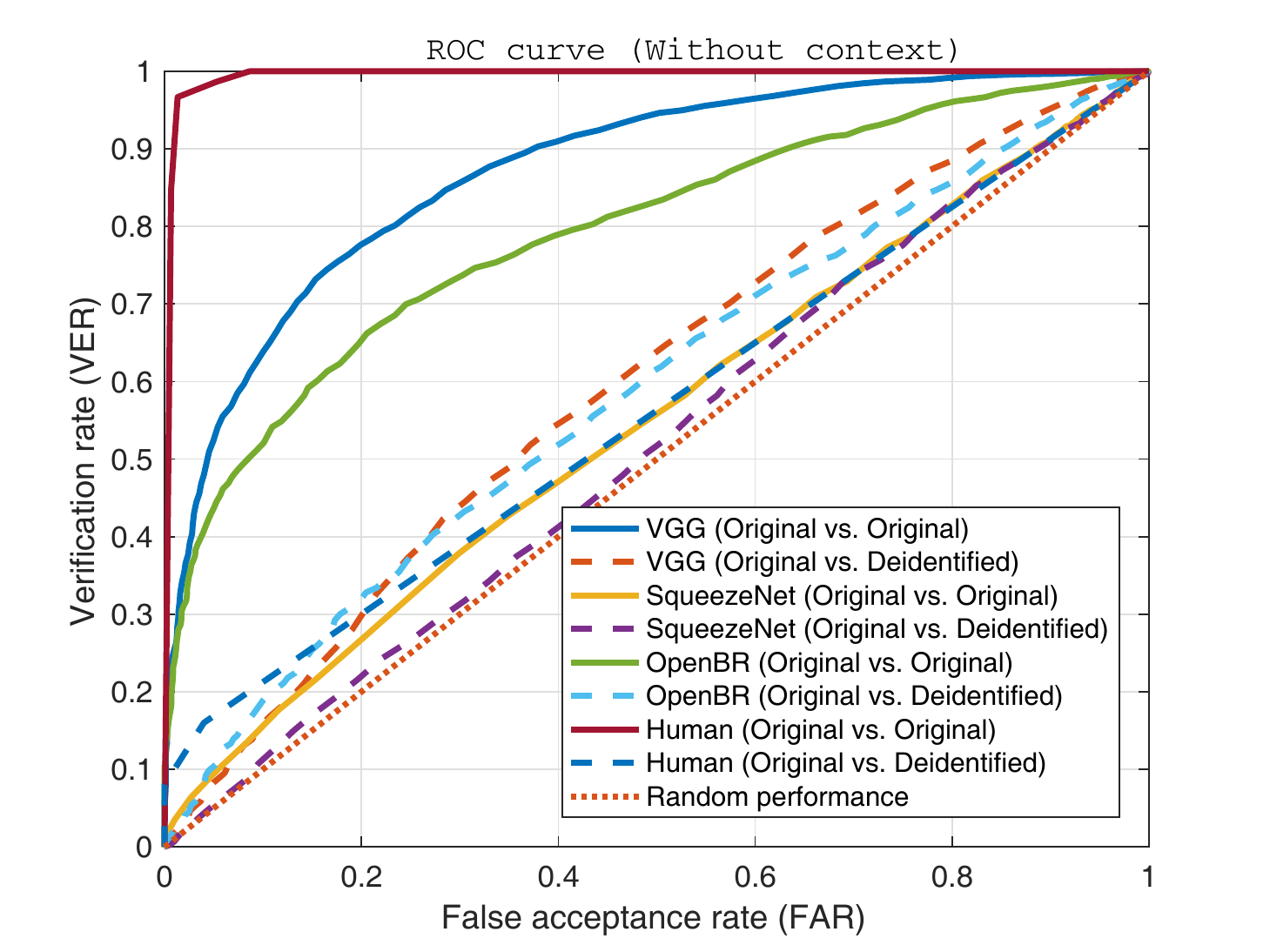}
\end{minipage}
\begin{minipage}{0.5\textwidth}
 \centering
 \includegraphics[width=1\textwidth]{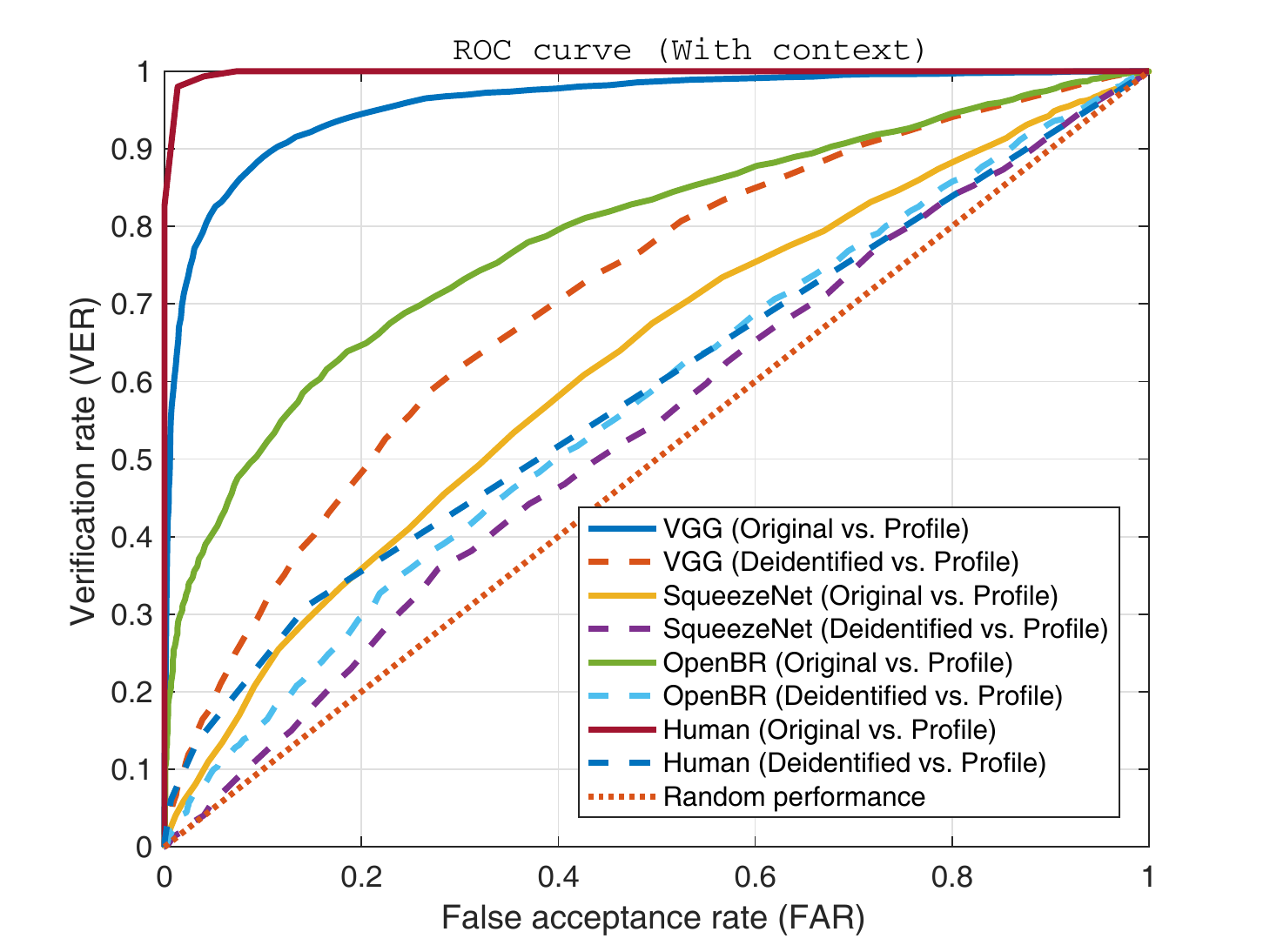}
\end{minipage}
\begin{minipage}{0.5\textwidth}
 \centering
 \includegraphics[width=1\textwidth]{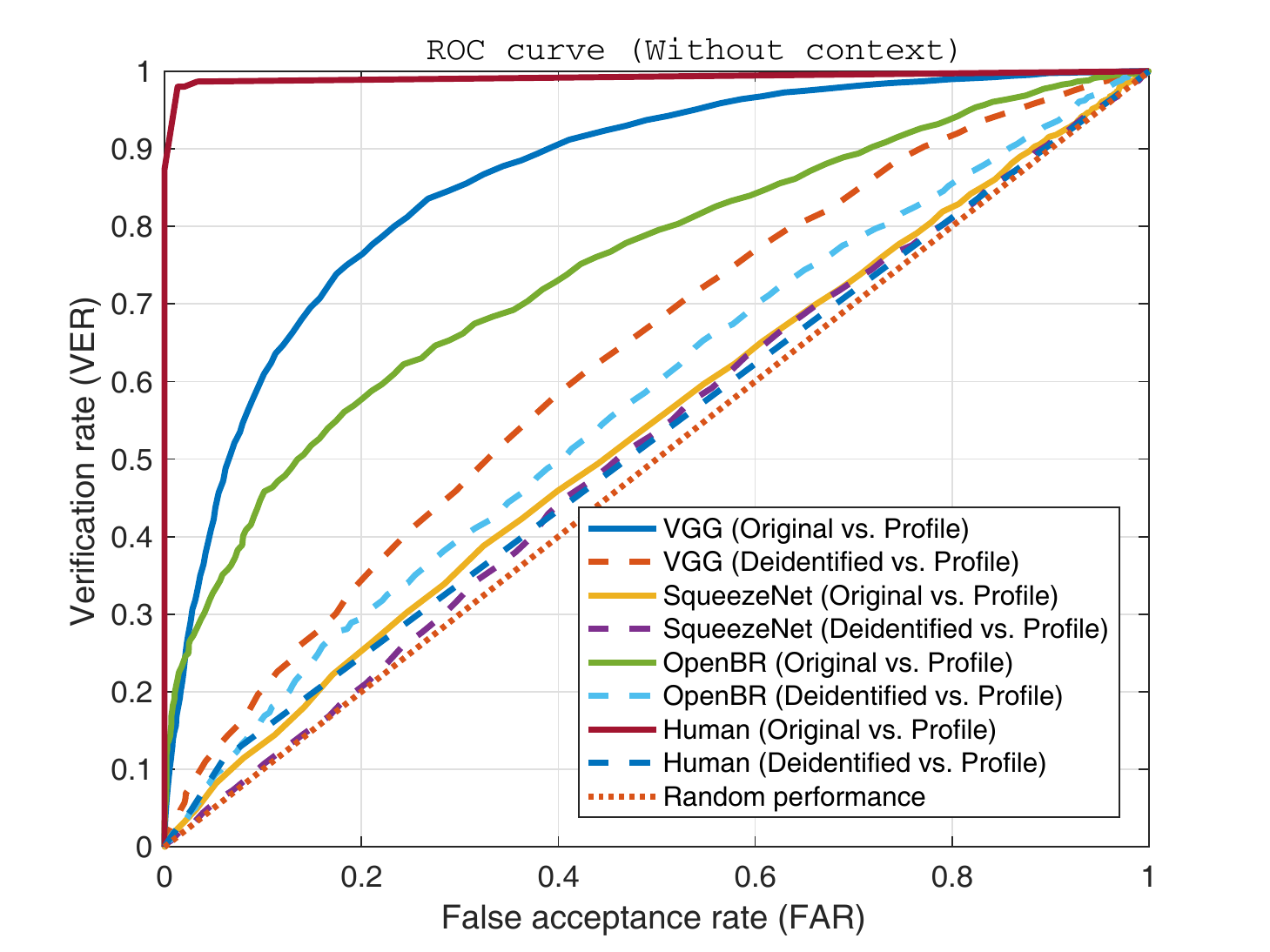}
\end{minipage}

\caption{ROC curves of the verification experiments. The curves on the left show the results of the experiments with images with contextual information and the curves on the right show the results obtained without contextual information. The upper row shows experiments with unaltered and deidentified images from the \textit{original} subset and the lower row shows experiments with unaltered and deidentified images from the \textit{profile} subset. All results show that our approach is effective.}
\label{fig:results_ROC}
\end{figure*}

\begin{table}[H]
\setlength{\tabcolsep}{3pt}
\renewcommand{\arraystretch}{1.2}
\caption{Quantitative results of the experiments. Average values and standard deviations over 10-fold are presented for all performance metrics. %The table shows results for all 4 types of verification experiments with images that include contextual information and images that do not. VER-1 values are not computable for manual (Human) experiments due to the limited number of verification attempt conducted. The results show that our approach is effective as the verification performance after deidentification is close to random in all experiments.
}
\footnotesize
\label{Tab:_numeric results}
\centering
\begin{tabular}{lrrrrrrrrrc}
\toprule
\multicolumn{2}{l}{{Test description}} & \multicolumn{2}{c}{{Original-to-original}} & \multicolumn{2}{c}{{Original-to-profile}} & \multicolumn{2}{c}{{Deidentified-to-original}}& \multicolumn{2}{c}{{Deidentified-to-profile}}& \multirow{2}{*}{Context illustration}\\ \cmidrule(l){1-10}% \midrule%\\
\multicolumn{2}{l}{{Metric (in \%)}}       & \multicolumn{1}{c}{EER}              & \multicolumn{1}{c}{VER-1}    & \multicolumn{1}{c}{EER}              & \multicolumn{1}{c}{VER-1}  & \multicolumn{1}{c}{EER}              & \multicolumn{1}{c}{VER-1}    & \multicolumn{1}{c}{EER}              & \multicolumn{1}{c}{VER-1} & \\\midrule
\multirow{4}{*}{\rotatebox[origin=c]{90}{Context}}    & VGG         & $8.7\pm1.0$  & $70.7\pm5.7$   &    $10.5\pm 1.3$  & $56.0\pm 10.3$& $34.4\pm2.2$& $4.2\pm3.0$   & $34.5\pm1.2$   & $5.5\pm3.2$  &\multirow{3}{*}{\vspace{8mm}\centering\includegraphics[width=2.6cm]{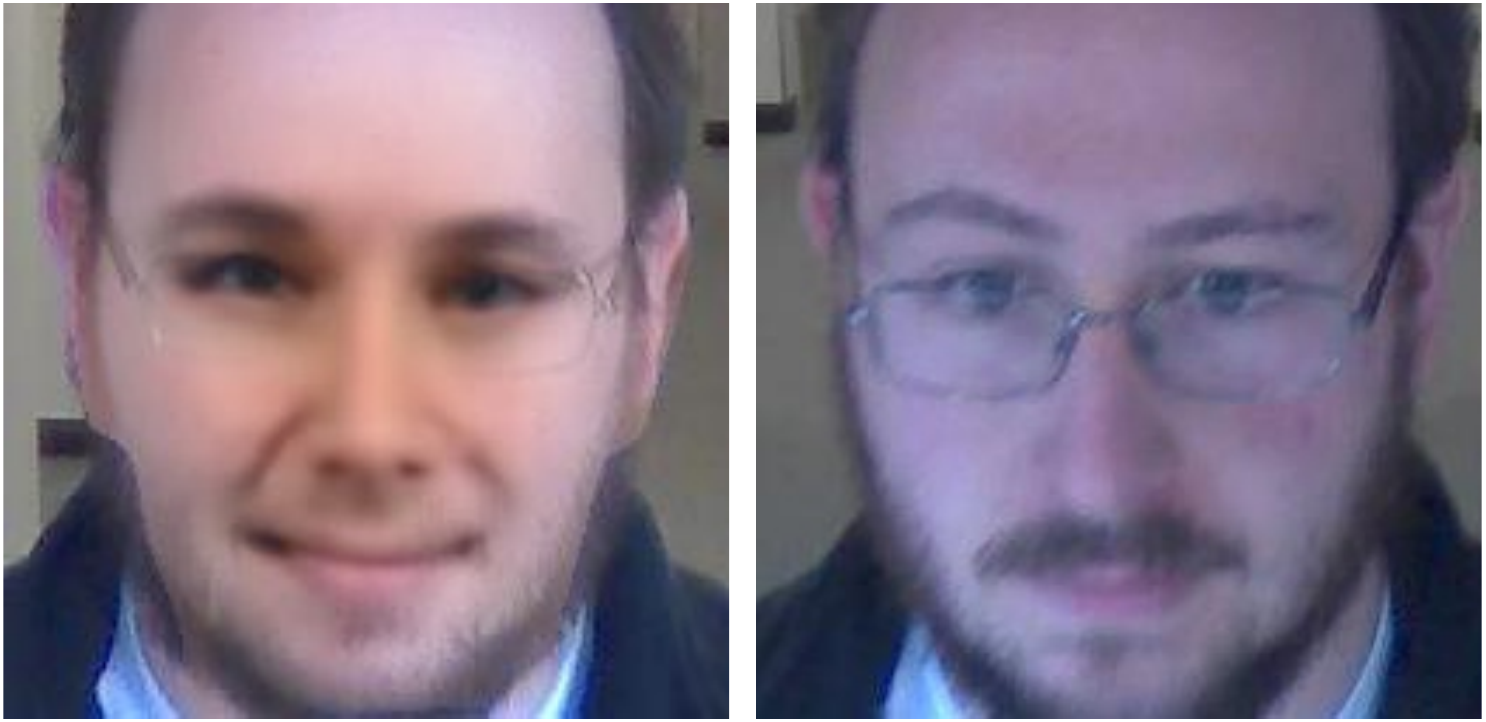}}\\
                                 & SqueezeNet  & $40.8\pm 2.8$  & $4.4\pm2.1$   &    $40.9\pm1.9$  & $3.3\pm1.6$& $47.3\pm2.3$& $0.8\pm0.7$   & $47.4\pm2.9$   & $1.2\pm1.0$  &\\
                                 & OpenBR      & $23.6\pm2.2$  & $34.5\pm6.8$   &    $28.3\pm1.7$  & $24.2\pm6.3$& $42.8\pm1.8$& $2.8\pm1.8$   & $45.3\pm2.8$   & $2.1\pm1.4$  &\\
                                 & Human       & $2.0\pm 2.8$  & $n/a$   &    $1.0\pm 2.2$  & $n/a$& $42.0\pm 7.6$& $n/a$   & $41.8\pm 7.2$   & $n/a$  &\\  \midrule
\multirow{4}{*}{\rotatebox[origin=c]{90}{No Context}} & VGG         & $21.5\pm2.9$  & $26.1\pm6.3$   &    $21.8\pm1.7$  & $13.1\pm5.7$& $43.3 \pm2.2$& $1.6\pm1.3$   & $40.6 \pm2.0$   & $3.4\pm1.5$  &\multirow{3}{*}{\vspace{8mm}\centering\includegraphics[width=2.6cm]{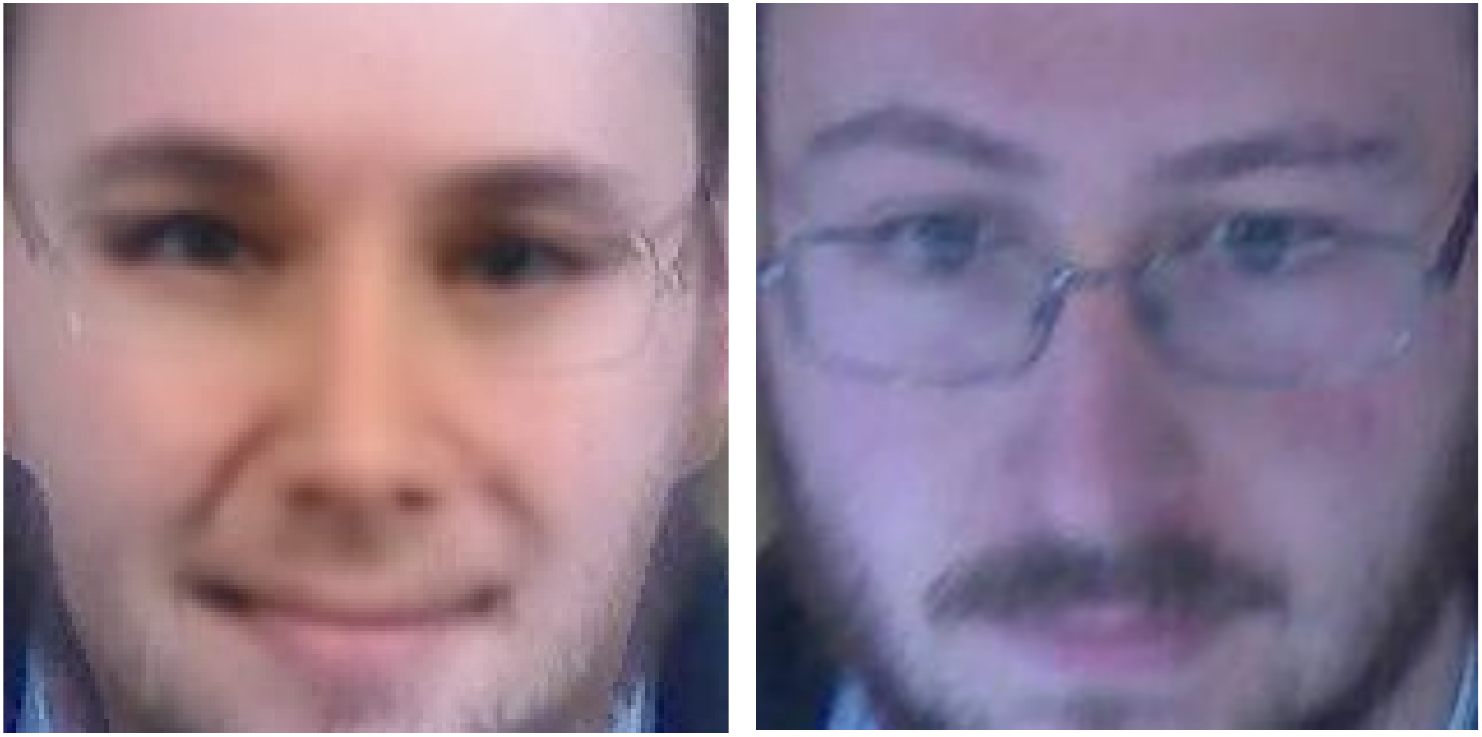}}\\
                                 & SqueezeNet  & $47.0\pm2.1$  & $3.8\pm1.3$   &    $47.1\pm2.1$  & $1.8\pm0.9$& $49.4\pm1.8$& $0.9\pm0.7$   & $48.4\pm2.1$   & $1.4\pm 1.3$  &\\
                                 & OpenBR      & $27.8\pm2.0$  & $19.1\pm6.6$   &    $32.2\pm3.1$  & $15.9\pm5.7$& $43.9\pm2.9$& $1.9\pm1.5$   & $45.2\pm2.9$   & $1.1\pm0.9$  &\\
                                 & Human       & $2.3\pm 3.2$  & $n/a$   &    $1.7\pm 2.8$  & $n/a$& $44.0\pm 8.4$& $n/a$   & $47.3\pm 5.8$   & $n/a$  &\\
\bottomrule
\end{tabular}
\end{table}
Numerical results of the experiments are presented in Table~\ref{Tab:_numeric results}. Note that VER-1 values are not reported for the manual experiments (denoted as Human) because of an insufficient number of manually graded image comparisons. As expected, the results with contextual information are significantly better than those without contextual information for all experiments when non-deidentified images are used. When the verification attempts are conducted with deidentified images, contextual information still contributes to a higher performance in all experiments, but the differences between images with and without context are smaller. As also evidenced by the ROC curves of the experiments in Fig.~\ref{fig:results_ROC}, the best performing automatic technique, the VGG network, is able to ensure a recognition performance well above random with an EER of 34.4\% for the \textit{deidentified-vs-original} experiment and an EER of 34.5\% for the \textit{deidentified-vs-profile} experiment when context is available. If no contextual information is present, the VGG performance drops to an EER of 43.3\% and 40.6\% for the same experiments, respectively. These observations suggest that contextual information is  important and may be exploited by contemporary recognition techniques to boost performance. Thus, care needs to be taken to appropriately conceal, modify or remove contextual information from the data as well.           

Another interesting observation that can be made from the ROC plots in Fig.~\ref{fig:results_ROC} is the drop in performance for the manual experiments. On the unaltered images, human performance is close to perfect for all experiments. However, after deidentification human performance drops to (more or less) random if no contextual information is present and is only slightly better than chance if contextual information is available. 
\begin{figure*}[htb]
 \centering
 \includegraphics[width=1\textwidth]{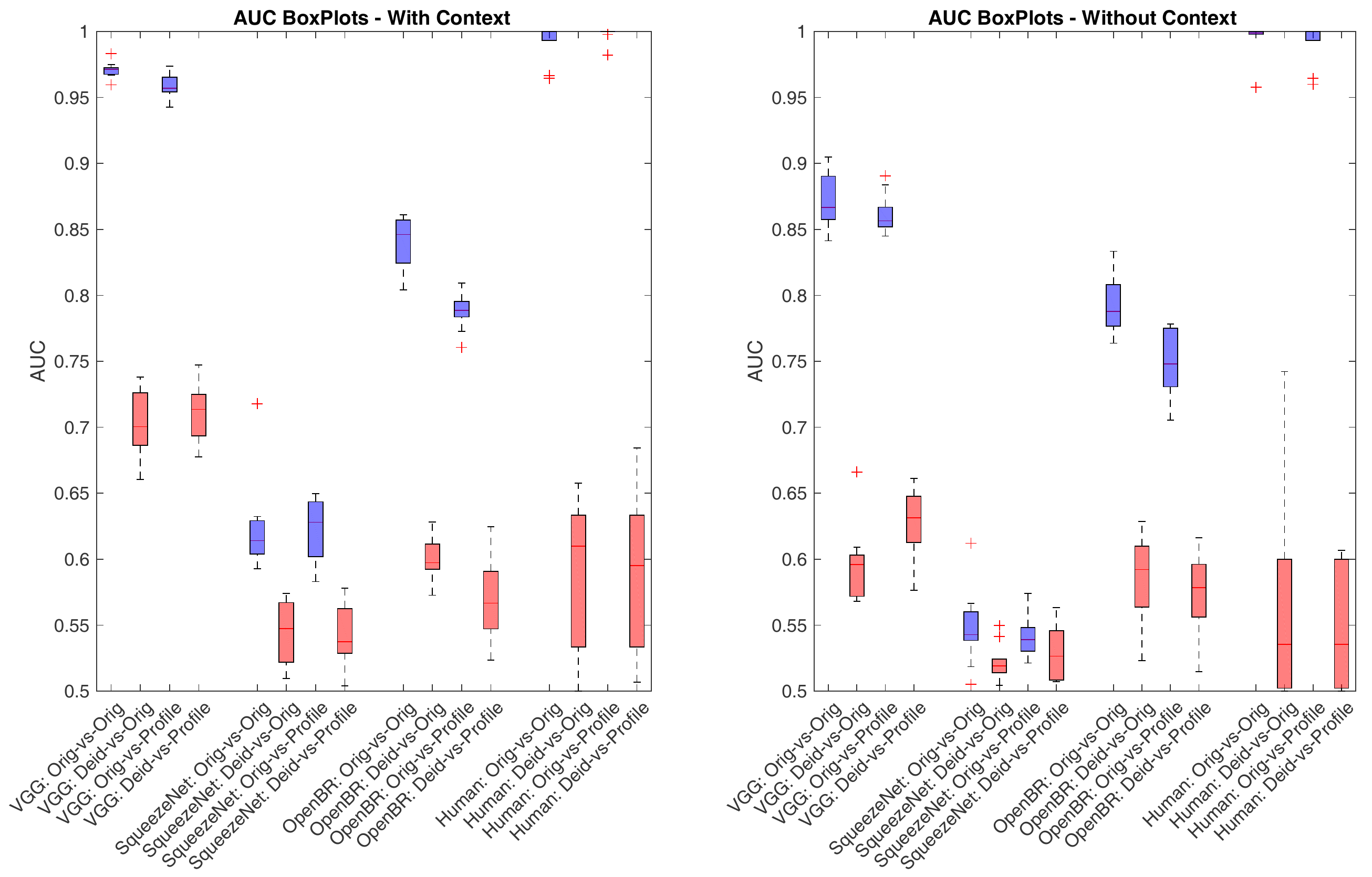}
\caption{AUC values from the 10 experimental folds presented in the form of box plots for all assessed techniques as well as human experiments. The blue  plots show results with unaltered images, the red plots show   experiments with deidentified images. The left box plots present the results with contextual information and the right box plots present the results without context. Note that after deidentification (red plots) the result are very close to 0.5 which indicated random performance.}
\label{fig:results_Boxplots}
\end{figure*}

The observations made so far are also supported by the  box-and-whiskers plots of the AUC values computed from the 10 experimental folds in Fig.~\ref{fig:results_Boxplots}. Here, a value of 0.5 indicates random performance. The blue box plots show results for unaltered images and the red box plots show the results for the deidentified images. %Each result with unaltered images (in blue) is followed directly by the corresponding experiment with an unaltered and deidentified image to enable direct comparisons. 
It needs to be noted that even in cases, when the performance after deidentification is not exactly random, it is still significantly lower than that obtained with unaltered images for all tested techniques. The highest median AUC value of any experiment after deidentification (AUC = 0.719) is achieved by the VGG network when contextual information is available. However, while this value is significantly above random it is of limited use to applications requiring reliable face recognition.  

% R1
%\textcolor{red}{
In our last experiment, we compare our deidentification approach to existing deidentification techniques from the literature. Specifically, we report results for two naive methods, i.e., blurring and pixelation, which unlike techniques from the $k$-Same family can be applied to video data using the same experimental protocol as used in the previous experiments. The results of the comparison are generated with the best performing recognition approach from Table~\ref{Tab:_numeric results}, i.e., the VGG network, and are presented in Table~\ref{Tab:comparison}. As can be seen, the naive methods result in worse recognition performance than our approach and therefore appear to ensure better anonymity. However, these methods destroy most of the information content of the images and can to a certain extent also be bypassed as shown by the results of the parrot (or imitation) attack experiments.
%}

%\textcolor{red}{
On the right side of Table~\ref{Tab:comparison} we show some qualitative deidentification examples on a closed set of images from the XM2VTS dataset \cite{Messer2003_XM2VTS} (top row). Note that a closed set is required for the $k$-Same family of techniques to be applicable. In accordance with the $k$-anonymity scheme \cite{Sweeney_Kanonym}, we replace clusters of (in this case $k=2$) images with the same surrogate face  generated by our GNN (the clusters are color-coded in the image). The results of our deidentification approach (last row) are visually convincing and feature no ghosting effects, such as the images generated by the original $k$-Same approach from \cite{Newton_original} (fourth row). With our approach it is also possible to retain certain aspects of the original data, which is not necessarily true for the blurred and pixelated images, shown in the second and third row of the image, respectively.
%}

\begin{table}[H]
\setlength{\tabcolsep}{3pt}
\renewcommand{\arraystretch}{1.3}
%\captionsetup{justification=raggedright}
\caption{
%\textcolor{red}{
Deidentification performance with the VGG network. The left part of the table shows a comparison of a few existing (naive) deidentification techniques and the proposed approach in experiments on the ChokePoint dataset. The right part of the table presents a qualitative comparison of our approach with competing techniques from the literature on a closed set of images. %The table shows results for all 4 types of verification experiments with images that include contextual information and images that do not. VER-1 values are not computable for manual (Human) experiments due to the limited number of verification attempt conducted. The results show that our approach is effective as the verification performance after deidentification is close to random in all experiments.
}
%}
\footnotesize
\label{Tab:comparison}
\centering
% R1
%\textcolor{red}{
\begin{tabular}{lrrrrr}
\toprule
\multirow{2}{*}{Deidentification technique \hspace{2mm}} & \multicolumn{2}{c}{{Context\hspace{2mm}}} & \multicolumn{2}{c}{{No Context}} & \hspace{4mm}Qualitative comparison (closed set)\hspace{4mm} \\ \cmidrule(l){2-6}
& EER (in \%) 	& VER-1 (in \%) &  EER (in \%) 	& VER-1 (in \%) & \hspace{4mm}\multirow{6}{*}{\vspace{8mm}\vspace{4mm}\centering\includegraphics[width=4.8cm]{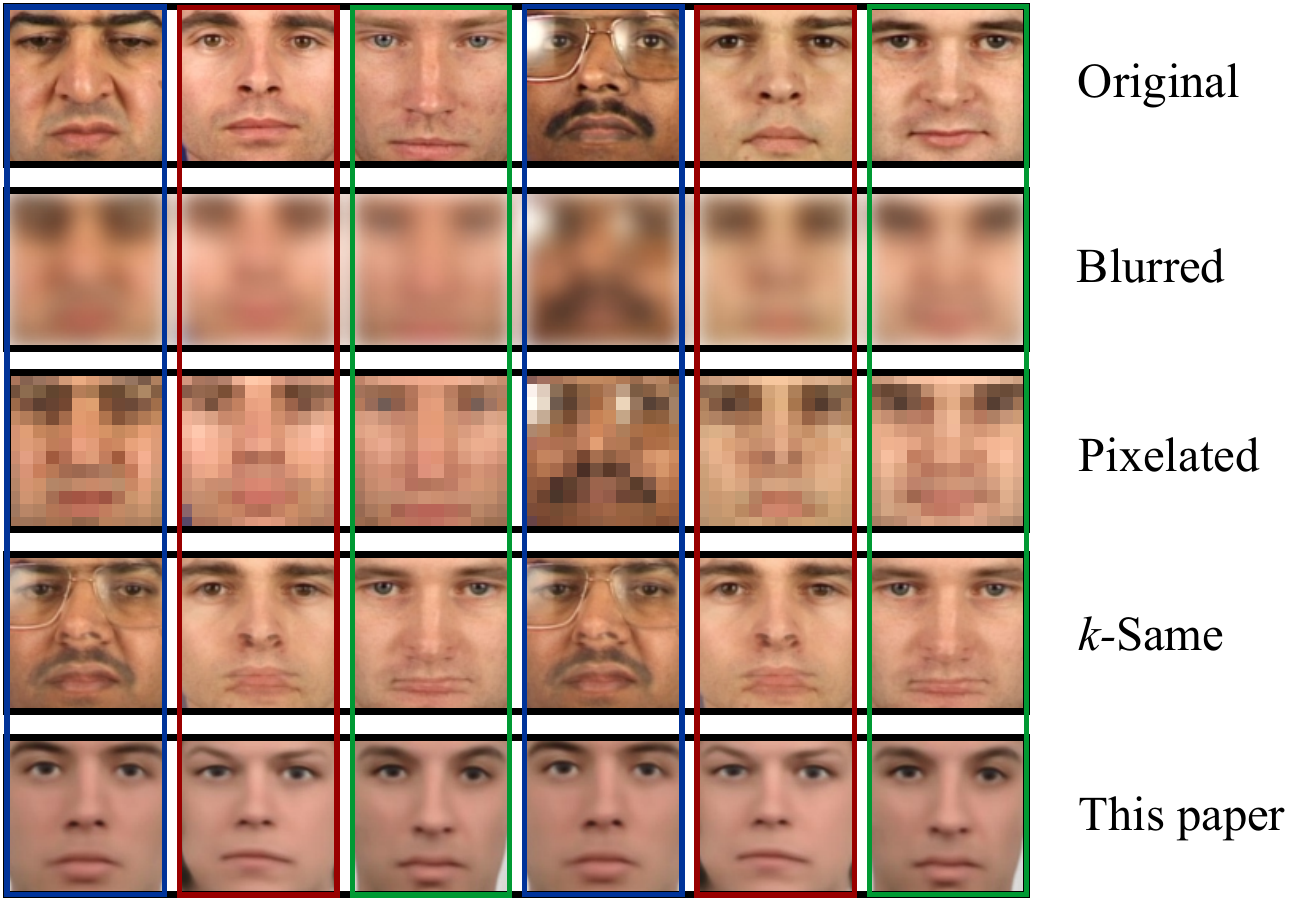}}\\
\cmidrule(l){1-5}
Pixelated				   	& $45.1\pm 1.8$			& $1.7\pm 0.9$			&	$47.3\pm 2.7$			& $1.3\pm 0.8$			&\\
Blurred						& $37.0\pm 1.4$			& $1.7\pm 1.2$			&  $43.0\pm 2.1$			& $1.5\pm 1.5$			&\\
Pixelated (parrot attack)	& $38.0\pm 1.4$			& $3.0\pm 1.6$			&  $39.4\pm 2.6$			& $2.7\pm 1.3$			&\\
Blurred (parrot attack)		& $32.4\pm 1.7$			& $13.9\pm 4.2$			&  $35.8\pm 2.1$			& $8.6\pm 3.4$			&\\
Ours						& $34.3\pm2.2$ 	& $4.2\pm3.0$	&  $43.2\pm2.2$	& $1.6\pm1.3$	&\\	
\cmidrule(l){1-5}
No deidentification			& $8.7\pm1.0$	& $70.7\pm5.7$ 	& $21.5\pm2.5$ 	& $26.1\pm6.3$ 	&\\
\bottomrule
\end{tabular}
%}
\end{table}

\section{Conclusion \label{sec:con}}
%Summary of what we did and presented, main findings and future work, 
In this paper we have presented a novel approach to face deidentification using generative neural networks. The proposed approach was evaluated on the ChokePoint dataset with highly encouraging results. Our evaluation suggests that generative networks are a viable tool for face deidentification and that a high degree of anonymity can be ensured by swapping the original faces by artificially generated surrogate faces. Furthermore, our experiments show that due to the flexibility of the generative network it is possible to control the appearance of the generated surrogate faces and thus retain (or alter) only specific aspects of the input images -- contributing significantly to the utility of the deidentified faces.  

% possible improvements
While our deidentification results are visibly convincing, additional improvements are possible. As part of our future work, we plan on 
including additional generator parameters to further capitalize on the utility of the deidentified faces. Other possible improvements include a better blending procedure that would improve the overall naturalness of the deidentified faces and remove artifacts. We will also consider incorporating a tracking scheme, which would improve the applicability of our approach on video data. % -- to exploit temporal information of processed data -- this would especially improve the cases where the detector misses a face or a face becomes partially occluded. %The latter case would also require to modify the replacement module to alter only the visible part of processed face, while keeping the background content intact.

\section*{Acknowledgement}

This research was supported in parts by the ARRS (Slovenian Research Agency) Research Programme P2-0250 (B) Metrology and Biometric Systems, the ARRS Research Programme P2-0214 (A) Computer Vision,  by TUBITAK project no. 113E067, by a Marie Curie FP7 Integration Grant within the 7th EU Framework Programme, by the Croatian HRRZ project DePPSS 6733 De-identification for Privacy Protection in Surveillance Systems and COST Action IC 1206 on De-identification for privacy protection in multimedia content.


\begin{thebibliography}{99}
%%%%%%%% sample bibliography
\bibitem{Ribaric_Review2016}
Ribari\'{c}, S., Ariyaeeinia, A., Pave\v{s}i\'{c}, N.: 'De-identification for privacy protection in multimedia content: A survey', Signal Processing: Image Communication, \textbf{47}, 2016, pp. 131--151.

\bibitem{PixelizationAndBlurring}
Neustaedter, C., Greenberg, S., Boyle, M.: 'Blur filtration
fails to preserve privacy for home-based video conferencing'.
%ACM 
TOCHI, 2005, pp. 1--36.

% newton, parrot attack
\bibitem{Newton_original} 
Newton, E.M., Sweeney, L., Malin, B.: 'Preserving privacy by de-identifying face images'. %IEEE 
TKDE, 2005, \textbf{17}, 2, pp. 232--243.

\bibitem{Gross_utility} 
Gross, R., Airoldi, E., Malin, B., Sweeney, L.: 'Integrating utility into face de-identification'. PET, 2005, 
%Springer Berlin Heidelberg, 
pp. 227--242.

\bibitem{Sim2015}
Sim, T., Zhang, L.: 'Controllable Face Privacy', AFGR, 2015, pp. 1--8.

% GNN references
\bibitem{Goodfellow_GAN2014}
Goodfellow, I.J., Pouget-Abadie, J., Mirza, M., Xu, B., Warde-Farley, D., Ozair, S., Courville, A., Bengio, Y.: 'Generative Adversarial Networks', arXiv:1406.2661, 2014.

\bibitem{Dosovitskiy_Chairs2015}
Dosovitskiy, A. and Sprigenberg, T. J., Brox T.:
'Learning to generate chairs with convolutional neural networks', CVPR, 2015, pp. 1538--1546.

\bibitem{VAE_GAN}
Larsen, A. B. L., Sonderby, S. K., Larochelle, H., Winther, O. : 'Autoencoding beyond pixels using a learned similarity metric', 
%arXiv preprint 
arXiv:1512.09300, 2015.

% chokepoint DB
\bibitem{Wong_Chokepoint2011}
Wong, Y., Chen, S., Mau, S., Sanderson, C., Lovell, B.C.: 'Patch-based Probabilistic Image Quality Assessment for Face Selection and Improved Video-based Face Recognition', CVPRW, 2011, pp. 81--88.
% Computer Vision and Pattern Recognition (CVPR)

\bibitem{Ribaric_Re}
Ribari\'{c}, S., Pave\v{s}i\'{c}, N.: 'An Overview of Face De-identification in Still Images and Videos'. 
%Int. Conf. and Works.  
%Automatic Face and Gesture Recognition (FG), 
%Ljubljana, Slovenia, May 2015, 
AFGR, 2015, pp. 1--6.

\bibitem{Sweeney_Kanonym} 
Sweeney, L.: '$k$-anonymity: a model for protecting privacy'. Uncertainty,
Fuzziness, and Knowledge-Based Systems, 2002, \textbf{10}, 5, pp. 557--570.

\bibitem{Machana_Ldiversity}
Machanavajjhala, A., Kifer, D., Gehrke, J., Venkitasubramaniam, M.: '$l$-diversity: Privacy beyond $k$-anonymity'. %ACM 
TKDD, 2007, \textbf{1}, 1, article 3.

\bibitem{Li_Tcloseness}
Li, N., Li, T., Venkatasubramanian, S.: '$t$-closeness: Privacy beyond $k$-anonymity and $l$-diversity'. 
%In Data Engineering, 2007. 
ICDE, 2007, pp. 106--115.
%(2007, April). 

% Active Appearance Model - AAM-based deid
\bibitem{Gross_MFM2008} 
Gross, R., Sweeney, L., de la Torre, F., Baker, S.: 'Semi-Supervised Learning of Multi-Factor Models for Face De-Identification', CVPR, 2008, pp. 1--8.

\bibitem{Gross_kSameM} 
Gross, R., Sweeney, L., de la Torre, F., Baker, S.: 'Model-Based Face De-Identification'. PRV, 2006, pp. 1--8.

\bibitem{Jourabloo_AAM2015}
Jourabloo, A., Yin, X., Liu, X.: 'Attribute Preserved Face De-identification', ICB, %May 
2015, pp. 278--285.

\bibitem{Samarzija_2014}
Samar\v{z}ija B., Ribari\'{c}, S.: 'An approach to the de-identification of faces in different poses'. 
%2014 37th International Convention on Information and Communication Technology, Electronics and Microelectronics (MIPRO), Opatija
MIPRO, 2014, pp. 1246--1251.

\bibitem{Sun2015}
Sun, Z., Meng, L., Ariyaeeinia A.: 'Distinguishable de-identified faces', AFGR, 2015, pp. 1--6.

% de-id workshop 2015
\bibitem{Brkic_ArtBased2016}
Brki\'{c}, K., Hrka\'{c}, T., Sikiri\'{c}, I., Kalafati\'{c}, Z.: 'Towards neural art-based face de-identification in video data', SPLINE, 2016, pp. 1--5. %July 2016

\bibitem{Chriskos2015}
Chriskos, P., Zoidi, O., Tefas, A., Pitas, I.: 'De-identifying facial images using projections on hyperspheres', AFGR, 2015, pp. 1--6.

\bibitem{Parkhi_Recog2015}
Parkhi, M. O., Vedaldi A., Zisserman A.: 'Deep Face Recognition', BMVC, 2015, article 41.%, p. 6. - cannot find the pages ... :-/ BMVC (Vol. 1, No. 3, p. 6) - 6 pages ?!
% British Machine Vision Conference

\bibitem{ViolaJones_Detector2001}
%Viola, P., Jones M.: 'Rapid object detection using a boosted cascade of simple features', CVPR, 2001. %, pp. I-I. -- are these pages?
Viola, P., Jones M.: 'Robust real-time face detection', IJCV, 2004, \textbf{57}, 2, pp. 137--154.

% Radbound Faces DB
\bibitem{Langner_RAFDB2010}
Langner, O., Dotsch, R., Bijlstra, G., Wigboldus, D.H.J., Hawk, S.T., van Knippenberg, A.: 'Presentation and validation of the Radboud Faces Database', Cognition\&Emotion, 2010, \textbf{24}, 8, pp. 1377--1388.

\bibitem{Kazemi_One2014}
Kazemi, V., Sullivan J.: 'One Millisecond Face Alignment with an Ensemble of Regression Trees', CVPR, 2014, pp. 1867--1874.
% Proc.

% ROC curves
\bibitem{ROC_curves}
Fawcett, T.: 'An introduction to ROC analysis'. PRL, 2006, \textbf{27}, 8, pp. 861--874.

\bibitem{Ziga_Hindawi}
Peer, P., Emer\v{s}i\v{c}, \v{Z}., Bule, J., \v{Z}ganec-Gros, J., \v{S}truc, V.: 'Strategies for exploiting independent cloud implementations of biometric experts in multibiometric scenarios'. MPE, \textbf{2014}, pp. 1--15.

\bibitem{Gajsek_ROC}
Gaj\v{s}ek, R., \v{S}truc, V., Dobri\v{s}ek, S., Miheli\v{c}, F.: 'Emotion recognition using linear transformations in combination with video'. Interspeech, 2009, pp. 1967--1970.

\bibitem{Neurocomputing}
Emer\v{s}i\v{c}, \v{Z}., \v{S}truc, V., Peer, P.: 'Ear Recognition: More than a Survey'. Neurocomputing, 2017, doi:10.1016/j.neucom.2016.08.139

% simson - NIST report
\bibitem{NIST_Simson}
Garfinkel, S.: 'De-Identification of Personal Information', 
%(NIST, October 2015), 
NISTIR 8053, 2015, pp. 1--46.

%squeezenet
\bibitem{SqueezeNet}
Iandola, F.N., Han, S., Moskewicz, M.W., Ashraf, K., Dally, W.J., Keutzer, K.: 'SqueezeNet: AlexNet-level accuracy with 50$\times$ fewer parameters and $< 0.5$ MB model size', 
%arXiv preprint 
arXiv:1602.07360, 2016.

% OpenBR
\bibitem{Klontz2013_OpenBR}
% R1
%\textcolor{red}{
Klontz, J., Klare, B., Klum, S., Jain, A., Burge., M.: 'Open Source Biometric Recognition'. BTAS, 2013, pp. 1--8.
%}

%human annotation
\bibitem{HumanAnot}
Phillips, J., O’Toole, A.: 'Comparison of human and computer performance across face recognition experiments', IVC, 2014, \textbf{32}, 1, pp. 74–-85.

% XM2VTS dataset
\bibitem{Messer2003_XM2VTS}
% R1
%\textcolor{red}{
Messer, K., Kittler, J., Sadeghi, M., Marcel, S., Marcel, C., Bengio, S., Cardinaux, F., Sanderson, C., Czyz, J., Vandendorpe, L., Srisuk, S., Petrou, M., Kurutach, W., Kadyrov, A., Paredes, R., Kepenekci, B., Tek, F.B., Akar, G.B., Deravi, F., Mavity, N.: 'Face Verification Competition on the XM2VTS Database'. AVBPA, 2003, pp. 964--974.
%}
%\bibitem{Du_GARP2014}
%Du, L., Yi, M., Blasch, E., Ling, H.: 'GARP-face: Balancing privacy protection and utility preservation in face %de-identification', IJCB, 2014, pp. 1--8.

%\bibitem{Radford_DCGAN2015}
%Radford, A., Metz, L., Chintala, S.: 'Unsupervised Representation Learning with Deep Convolutional Generative Adversarial Networks', arXiv:1511.06434, 2015.

%\bibitem{Salimans_ImprGANs2016}
%Tim Salimans, Ian Goodfellow, Wojciech Zaremba, Vicki Cheung, Alec Radford, Xi Chen: 'Improved Techniques for Training GANs'

%\bibitem{Dosovitskiy_Chairs2014}
%Dosovitskiy, A. and Sprigenberg, T. J. and Brox T.: 'Learning to Generate Chairs with Convolutional Neural Networks', arXiv, 2014 % TODO


%\bibitem{ROC_evalRoss}
%Beveridge, J. R., Hao Z., Patrick J. F., Lee, Y., Liong, V.E., Lu, J., Assis Angeloni, M. de, Freitas Pereira, T. de, Li. H., Hua, G., Struc, V., Krizaj, J., Phillips, J.: 'The IJCB 2014 PaSC video face and person recognition competition'. IJCB, 2014, pp. 1-8. 


%Conf. Systems Biology, Stockholm, Sweden, May 2006, pp. 1--7
%\vbox{\subsection{Websites}}
%
%\bibitem{bib1}
%
%`Author Guide - IET Research Journals', http://digital-library.theiet.org/journals/author-guide, accessed 27
%November 2014
%
%\bibitem{bib2}
%`Research journal length policy', http://digital-library.theiet.org/files/research\_journals\_\break length\_policy.pdf, accessed 27
%November 2014
%
%\bibitem{bib3}
%`ORCID: Connecting research and researchers', http://orcid.org/, accessed 3 December 2014
%
%\bibitem{bib4}
%`Fundref', http://www.crossref.org/fundref/, accessed 4 December 2014
%
%\vbox{\subsection{Journal articles}}
%
%\bibitem{bib5}
%Smith, T., Jones, M.: 'The title of the paper', IET Syst. Biol., 2007, \textbf{1}, (2), pp. 1--7
%
%\bibitem{bib6}
%Borwn, L., Thomas, H., James, C.,~\textit{et al}.:'The title of the paper, IET
%Communications, 2012, \textbf{6}, (5), pp 125--138
%
%\vbox{\subsection{Conference Paper}}
%
%\bibitem{bib7}
%Jones, L., Brown, D.: 'The title of the conference paper'. Proc. Int.
%Conf. Systems Biology, Stockholm, Sweden, May 2006, pp. 1--7
%
%\vbox{\subsection{Book, book chapter and manual}}
%
%\bibitem{bib8}
%Hodges, A., Smith, N.: 'The title of the book chapter', in Brown, S.
%(Ed.): 'Handbook of Systems Biology' (IEE Press, 2004, 1st edn.), pp. 1--7
%
%\bibitem{bib9}
%Harrison, E.A., and Abbott, C.: 'The title of the book' (XYZ Press,
%2005, 2nd edn. 2006)
%
%\vbox{\subsection{Report}}
%
%\bibitem{bib10}
%IET., 'Report Title' (Publisher, 2013), pp. 1-5
%
%\vbox{\subsection{Patent}}
%
%\bibitem{bib11}
%Brown, F.: 'The title of the patent (if available)'. British Patent
%123456, July 2004
%
%\bibitem{bib12}
%Smith, D., Hodges, J.: British Patent Application 98765, 1925
%
%\vbox{\subsection{Thesis}}
%
%\bibitem{bib13}
%Abbott, N.L.: 'The title of the thesis'. PhD thesis, XYZ University, 2005
%
%\vbox{\subsection{Standard}}
%
%\bibitem{bib14}
%BS1234: 'The title of the standard', 2006

\end{thebibliography}
\end{document}